%% file: main.tex
\documentclass[runningheads]{llncs}
\input{setup/package}
\input{setup/macros}

\input{setup/symbols}
\input{setup/graphicspath}

\usepackage{eccv}

\usepackage{eccvabbrv}

\usepackage{graphicx}
\usepackage{booktabs}
\usepackage{algorithmic}

\usepackage[accsupp]{axessibility}  

\usepackage{hyperref}

\usepackage{orcidlink}

\begin{document}

\title{DEVIAS: Learning Disentangled Video Representations of Action and Scene} 

\titlerunning{DEVIAS: Learning Disentangled Video Representations of Action and Scene}

\author{Kyungho Bae\inst{1*} \and
Geo Ahn\inst{1*}  \and
Youngrae Kim\inst{2*} \and
Jinwoo Choi\inst{1\dagger} 
}


\authorrunning{K. Bae \etal}


\institute{%
${}^{1}$Kyung Hee University \quad ${}^{2}$KAIST \\
\email{\{kyungho.bae, ahngeo11, jinwoochoi\}@khu.ac.kr}, \email{youngrae.kim@kaist.ac.kr}}

\maketitle
\let\thefootnote\relax\footnote{${}^{*}$Equally contributed first authors. ${}^{\dagger}$Corresponding author.}

\input{0_abstract}
\input{1_introduction}
\input{2_prior}

\input{4_method}
\input{5_result}

\input{6_conclusion}

\paragraph{Acknowledgment.}
This work was supported in part by the Institute of Information \& Communications Technology Planning \& Evaluation (IITP) grant funded by the Korea Government (MSIT) under grant RS-2024-00353131, RS-2021-II212068 (Artificial Intelligence Innovation Hub), and RS-2022-00155911 (Artificial Intelligence Convergence Innovation Human Resources Development
(Kyung Hee University)).
Additionally, it was supported by the National Research Foundation of Korea (NRF) grant funded by the Korea Government (MSIT) (No. 2022R1F1A1070997).
{\small
\bibliographystyle{splncs04}
\bibliography{main}
}

\clearpage
\begin{center}
{\Large\bfseries DEVIAS: Learning Disentangled Video Representations of Action and Scene}
\vspace{0.5cm}  

{\large Supplementary Material}
\end{center}

\vspace{1cm}  

\input{99_supp_content}

\end{document}

%% file: setup/package.tex
\usepackage{graphicx}
\usepackage{subcaption}
\usepackage{float}
\usepackage[justification=justified]{caption}	
\usepackage{lscape}                                         

\usepackage[lined,ruled,linesnumbered]{algorithm2e}

\usepackage{booktabs}                   
\usepackage{multirow}
\usepackage{makecell}

\usepackage{paralist}
\usepackage{enumitem}

\usepackage{bm}                          
\usepackage{epsfig}                      
\usepackage{graphicx}                  
\usepackage{mathtools}
\usepackage{amssymb}   
\usepackage{amsmath}
\usepackage{amsfonts}
\usepackage{bbm}

\usepackage{units}
\usepackage{color}

\usepackage{comment}
\usepackage{setspace}

\usepackage{url}  
\usepackage[pagebackref=true,breaklinks=true,letterpaper=true,colorlinks,bookmarks=false]{hyperref}

\usepackage[capitalize]{cleveref}
\crefname{section}{Sec.}{Secs.}
\Crefname{section}{Section}{Sections}
\Crefname{table}{Table}{Tables}
\crefname{table}{Tab.}{Tabs.}

\usepackage{xspace}
\usepackage{colortbl}
\usepackage{setspace}

\usepackage{wrapfig}

%% file: setup/macros.tex
\def\etal{et~al.\_}			  
\def\eg{e.g.,~}               
\def\ie{i.e.,~}               
\def\vs{vs.~}                 

\newlength\paramarginsize
\newlength\figmarginsize
\newlength\tabmarginsize
\newlength\secmarginsize
\newlength\figcapmarginsize
\newlength\tabcapmarginsize

\newcommand{\figcapmargin}{\vspace{\figcapmarginsize}}

\newcommand{\mpage}[2]
{
\begin{minipage}{#1\linewidth}\centering
#2
\end{minipage}
}

\newcommand{\secref}[1]{Section~\ref{sec:#1}}
\newcommand{\figref}[1]{Figure~\ref{fig:#1}} 
\newcommand{\tabref}[1]{Table~\ref{tab:#1}}
\newcommand{\eqnref}[1]{\eqref{eq:#1}}

\newcommand{\algref}[1]{Algorithm~\ref{#1}}

\long\def\ignorethis#1{}

\newcommand{\tb}[1]{\textbf{#1}}

\newcommand{\best}[1]{{\textbf{#1}}}
\newcommand{\second}[1]{{\underline{#1}}}

\newcommand{\mlp}{\texttt{MLP}}

%% file: setup/symbols.tex
\def\xi{\mathbf{x}_i}

\def\A{\mathbf{A}}

\def\H{\mathbf{H}}
\def\S{\mathbf{S}}

\def\Q{\mathbf{Q}}
\def\K{\mathbf{K}}
\def\V{\mathbf{V}}
\def\M{\mathbf{M}}
\def\X{\mathbf{X}}

\def\Z{\mathbf{Z}}
\def\W{\mathbf{W}}

%% file: setup/graphicspath.tex
\graphicspath{{figure}, {images}, {example}}

%% file: 0_abstract.tex
\begin{abstract}

Video recognition models often learn scene-biased action representation due to the spurious correlation between actions and scenes in the training data.
Such models show poor performance when the test data consists of videos with unseen action-scene combinations. 
%
%
Although scene-debiased action recognition models might address the issue, they often overlook valuable scene information in the data. 
To address this challenge, we propose to learn DisEntangled VIdeo representations of Action and Scene (DEVIAS), for more holistic video understanding. 
%
%
We propose an encoder-decoder architecture to learn disentangled action and scene representations with a single model.
The architecture consists of a disentangling encoder (DE), an action mask decoder (AMD), and a prediction head.
The key to achieving the disentanglement is employing both DE and AMD during training time.
%
%
The DE uses the slot attention mechanism to learn disentangled action and scene representations. 
%
%
For further disentanglement, an AMD learns to predict action masks, given an action slot.
%
With the resulting disentangled representations, we can achieve robust performance across diverse scenarios, including both seen and unseen action-scene combinations.
%
We rigorously validate the proposed method on the UCF-101, Kinetics-400, and HVU datasets for the seen, and the SCUBA, HAT, and HVU datasets for unseen action-scene combination scenarios.
%
%
Furthermore, DEVIAS provides flexibility to adjust the emphasis on action or scene information depending on dataset characteristics for downstream tasks. 
DEVIAS shows favorable performance in various downstream tasks: Diving48, Something-Something-V2, UCF-101, and ActivityNet. The code is available at \href{https://github.com/KHU-VLL/DEVIAS}{https://github.com/KHU-VLL/DEVIAS}.

%
%
%

\keywords{Action recognition \and Video recognition \and Scene recognition \and Video representation learning \and Disentangled representation learning}

\end{abstract}

%% file: 1_introduction.tex
\section{Introduction}
\label{sec:intro}

Humans can naturally understand the content of a video by extracting human actions from the surrounding scene context.
Even when encountering a previously unseen action-scene combination, humans easily recognize both the action and the scene: \eg in \figref{teaser} (b), the people are \emph{dancing} in a \emph{football field}. 

Unlike humans, most video action recognition methods struggle to decompose actions and scene context from an input video.
Instead, video action recognition methods tend to learn scene-biased action representation due to the spurious correlation between actions and scenes in the video datasets.
The existing video datasets~\cite{soomro2012ucf101,kay2017kinetics,caba2015activitynet} often consist of limited combinations of action-scene pairs for each action class, \eg if the scene of a video is a football field, the action is always \emph{playing football} and vice versa.
However, in reality, diverse actions such as dancing or cheerleading can also take place on a football field~\cite{whycantchoi}. 
The reason for the limited action-scene combinations in the dataset stems from the high cost of constructing a video dataset with diverse combinations, rather than a lack of such scenarios in the real world. 
Therefore, a desired model should have robust performance across diverse action-scene combination scenarios.

\input{figure/fig_teaser}
A scene-biased action recognition model is likely to predict actions based on the scene context rather than the action itself, leading to errors when encountering unseen action-scene combinations~\cite{li2018resound,whycantchoi,chung2022hatdataset}.
For example, as shown in \figref{teaser} (b), scene-biased action recognition models are likely to misclassify the action as \emph{playing football} instead of \emph{dancing}.
Scene-debiased action recognition models~\cite{whycantchoi,fame,BE,li2023stillmix} might be a solution to the problem.
The scene-debiased action recognition models show significant improvement for unseen action-scene combinations~\cite{li2023stillmix} and for scenarios where actions and scenes are barely correlated~\cite{li2018resound,mimetics}.
However, as illustrated in \figref{teaser}, scene-debiased action recognition models often overlook the scene context as they are trained to disregard scene information, omitting potentially valuable context.

In this work, we move beyond the limitations of prior works that disregard scene context.
We tackle an interesting yet relatively under-explored problem: learning DisEntangled VIdeo representations of Action and Scene (DEVIAS) for holistic video understanding.
%
%
Having \emph{both action and scene} representations provides richer information than scene-debiased representations.
With \emph{disentangled} action and scene representations, a video model can understand video regardless of seen or unseen action-scene combinations.
As illustrated in the third row of \figref{teaser}, with disentangled representations, a model could accurately recognize that the people are \emph{dancing} whether on a \emph{stage} (a) or in a \emph{football field} (b).

Disentangled action-scene representations allow for tailored applications; one can adjust the emphasis on action or scene to suit specific tasks.
For instance, we can leverage scene context in some datasets where it is beneficial~\cite{soomro2012ucf101,kay2017kinetics,caba2015activitynet} to boost action performance.
On the other hand, in some tasks where scene information is non-beneficial, we could encourage a model to focus on the action rather than the scene context: \eg barely correlated actions and scenes~\cite{goyal2017something,li2018resound,mimetics}, and action-scene combinations vary between training and test phases~\cite{miech20rareact,chung2022hatdataset,li2023stillmix}.


In this paper, we propose DEVIAS, a novel encoder-decoder architecture to learn disentangled representations of action and scene with a single model. 
%
DEVIAS consists of three parts: a disentangling encoder (DE), an action mask decoder (AMD), and an action/scene classification head.
%
The key to achieving the disentanglement is employing \emph{both DE and AMD during training time}.
%
In the DE, we employ the slot attention~\cite{slot_object_centric} to learn disentangled action and scene representations.
In the slot attention, multiple learnable slots compete with each other as we normalize the attention coefficients over the slot axis.
After a few iterations of slot attention, the learnable slots progressively learn distinct information, \eg action, and scene.
On top of DE, AMD further disentangles action and scene representations.
AMD is a lightweight decoder that learns to predict action masks given an action slot as an input.
Thanks to the complementary nature of the slot attention, DEVIAS encourages the DE to learn not only good action representation but also good scene representation.
As a result, the DE effectively learns disentangled action and scene representations.

To validate the effectiveness of DEVIAS, we carefully design a set of controlled experiments.
Through the experiments, we verify the effectiveness of each representation in seen action-scene combination scenarios: UCF-101~\cite{soomro2012ucf101}, Kinetics-400~\cite{kay2017kinetics}, and HVU~\cite{diba2020hvu} and in unseen action-scene combination scenarios: SCUBA~\cite{li2023stillmix}, HAT~\cite{chung2022hatdataset}, and HVU~\cite{diba2020hvu}.
DEVIAS shows favorable performance over the baselines in both seen and unseen combination scenarios. 
DEVIAS also show favorable performance on various downstream tasks: Diving48~\cite{li2018resound}, Something-Something-V2~\cite{goyal2017something}, UCF-101~\cite{soomro2012ucf101}, and ActivityNet~\cite{caba2015activitynet}.

In this work, we make the following major contributions:
\begin{itemize}
\item We tackle an interesting and challenging yet relatively under-explored problem of learning \emph{disentangled action and scene representations}. 
%
We aim to \emph{shift the paradigm} from merely recognizing actions to recognizing both actions and scenes in the video recognition field.
%
\item We introduce DEVIAS, a novel encoder-decoder architecture designed to learn disentangled action and scene representations. 
The key to achieving the disentanglement is employing both the disentangling encoder with slot attention and the action mask decoder during training time.
%

\item We conduct extensive experiments to validate the effectiveness of DEVIAS. 
%
DEVIAS shows robust performance over the baselines, in both seen and unseen action-scene combination scenarios and on various downstream tasks.
\end{itemize}

%% file: figure/fig_teaser.tex
\begin{figure*}[t]
    \mpage{0.48}{
        \centering
        \includegraphics[width=\linewidth]{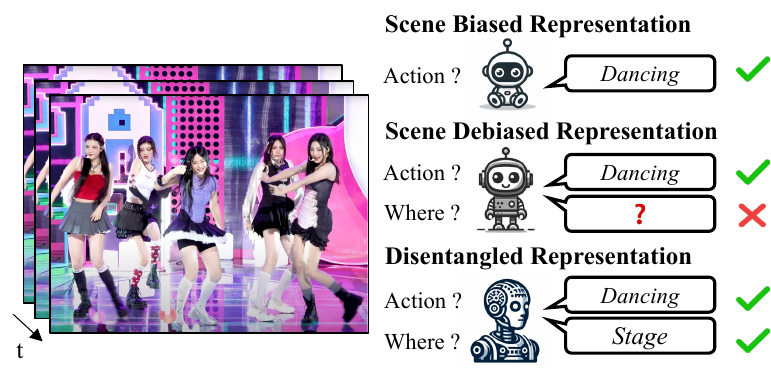}
    }
    \hfill
    \mpage{0.48}{
        \centering
        \includegraphics[width=\linewidth]{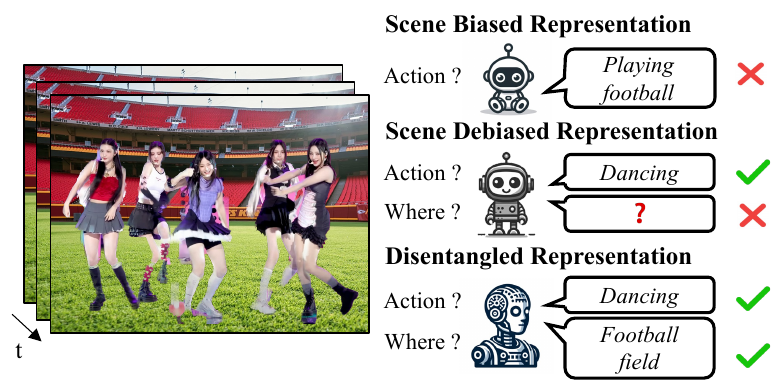}
    }

    \figcapmargin
    \mpage{0.48}{\fontsize{5pt}{10pt}\selectfont (a) Seen combination: \emph{dancing} on a \emph{stage}.}\hfill
    \mpage{0.48}{\fontsize{5pt}{10pt}\selectfont (b) Unseen combination: \emph{dancing} in a \emph{football field}.}

    \vspace{-0.2em} 
    \captionof{figure}{
        %
        %
        \textbf{Why do we need disentangled video representations?}
        Video recognition models often learn scene-biased representations due to the spurious correlation~\cite{li2018resound,whycantchoi} between action and scene in the dataset. Although such models might work well when video contains an action-scene combination seen during training \eg (a), they would fail when video contains an unseen combination \eg (b). In contrast, scene debiased video models~\cite{BE,fame} could accurately recognize the action regardless of combinations. However, they are not capable of predicting scenes. In this work, we propose to learn \emph{disentangled} action and scene representations. The disentangled model can understand both action and scene well, including both seen and unseen action-scene combinations, \eg it correctly predicts \emph{dancing} on a \emph{stage} (a) and  \emph{dancing} in a \emph{football field} (b).        
    }
    \figcapmargin  
    \label{fig:teaser}
\end{figure*}

%% file: 2_prior.tex
\section{Related Work}
\label{sec:related}
\noindent\textbf{Video action recognition.} 
The last decade has seen remarkable progress in video action recognition.
Key approaches to action recognition include 2D CNNs~\cite{donahue2016longterm,karpathy20142dcnn,lin2019tsm,Simonyan-NIPS-2014,ng2015short,zhou2018trn}, two-stream CNNs~\cite{Simonyan-NIPS-2014,feichtenhofer2019slowfast}, 3D CNNs~\cite{carreira2018i3d,feichtenhofer2019slowfast,ji20123d,tran2015c3d,tran2018closer,wang2018non,feichtenhofer2020x3d}, 2D and 1D separable CNNs~\cite{tran2018closer, Xie-ECCV-2018} or more recent transformer architectures~\cite{arnab2021vivit, timesformer, herzig2022object, motionformer, wu2022memvit, fan2021multiscale, yan2022multiview,girdhar2022omnivore,standalone}. 
Despite the great progress, a common limitation among the previous works is scene-biased action representation.
The bias stems from the spurious correlation between action and scene in the training data, often leading to poor performance on the test data with different distributions. 
In contrast to the prior works, we aim to reduce the influence of the spurious correlation by learning disentangled representations of action and scene.


\noindent\textbf{Mitigating scene bias in action recognition.}
The community has identified scene bias~\cite{li2018resound,whycantchoi} as the devil of action recognition as biased models do not generalize well to new tasks/domains.
Scene debiasing is beneficial, enhancing performance in downstream tasks~\cite{whycantchoi,BE,fame,ding2022dual}, data-efficiency~\cite{actorcut,gowda2022learn2augment}, and domain adaptation~\cite{choi2020sava,sahoo2021comix}.
Yet, an often-overlooked aspect is that action representation, when debiased from the scene context, may lose valuable contextual information. 
In contrast to scene-debiasing, DEVIAS learns disentangled action and scene representations.
The disentangled representations enable tailored emphasis on either action or scene depending on the specific downstream tasks and datasets.
%


\noindent\textbf{Disentangled representation learning.}
Generative modeling works have extensively explored disentangled representation learning to manipulate each attribute for image and video generation~\cite{dcign,chen2016infogan,betavae,bhagat2020disentangling,denton2017unsupervised,massague2020learning,wang2019self,qian2022static,wang2020g3an,hsieh2018learning,lai2021video,xing2020deformable}. 
%
Disentangled representation learning for video understanding includes disentangling different attributes~\cite{tran2017disentangled,Zhang_2019_CVPR}, learning dynamic and static components of videos~\cite{lstm-decompose,zhao2018recognize}, and disentangling action and scene~\cite{wang2018pulling,cvprworkshop-disent} to improve action recognition.
We also focus on disentangled representations for video understanding.
Unlike prior works that overlook the quality and utility of the scene representation~\cite{wang2018pulling,cvprworkshop-disent}, we aim to learn not only high-quality action but also high-quality scene representation.
%
To the best of our knowledge, DEVIAS is a pioneering effort in achieving this balanced focus on both action and scene recognition.

%% file: 4_method.tex
\section{DEVIAS}
\label{sec:method}

\input{figure/fig_overview}

We introduce DEVIAS, an encoder-decoder architecture designed to learn disentangled action and scene representations within a unified model, as illustrated in \figref{overview}.
DEVIAS consists of i) a disentangling encoder (DE), ii) an action mask decoder (AMD), and iii) an action/scene prediction head.
To learn disentangled representations, it is crucial to employ \emph{both DE and AMD} during training.


Given an input video, DE extracts a feature vector $\X\in\mathbb{R}^{NT \times D}$ using a backbone transformer encoder, where $N$ is the number of spatial patches, $T$ is the number of frames, and $D$ denotes the dimension of patch embeddings.
DE captures distinct information from the feature vectors, \ie action, and scene by slot attention.
Given $K$ learnable slots as queries, the slot attention iteratively attends to encoded features as keys and values. 
Then the DE outputs disentangled slots.
%
With the slot attention, DEVIAS learns action and scene representations by competition among multiple slots.
Then a matching function solves $K$ to $2$ matching problem to determine an action and a scene slot.
We supervise action slot learning using action labels covering $N_A$ action types and scene slot learning using scene labels covering $N_S$ scene types. 
We use a shared head for the action and scene class prediction.
For disentanglement, we employ a lightweight action mask decoder. 
Given an action slot, AMD learns to predict action masks.
We provide detailed descriptions of the DE and the prediction head in \secref{disentangle}, AMD in \secref{aux}, and training \& inference of DEVIAS in \secref{tr_inf}.

\subsection{Disentangling Encoder}
\label{sec:disentangle}

\noindent\textbf{Slot attention for disentangled representation learning.}
To learn disentangled action and scene representations, DE employs the slot attention mechanism~\cite{slot_object_centric,setmodule}.
In each slot attention iteration, we project an input feature vector $\X\in\mathbb{R}^{NT \times D}$, and $K$ learnable slots $\S\in\mathbb{R}^{K \times D}$ to a common space with the dimension $D_h$ as follows: 
$\Q = \S \W_Q \in \mathbb{R}^{K \times D_h}$, 
$\K = \X \W_K \in \mathbb{R}^{NT \times D_h}$, and 
$\V = \X \W_V \in \mathbb{R}^{NT \times D_h}$, 
where $\W_{Q}$, $\W_{K}$, and $\W_{V}$ are the $D \times D_h$ query, key, and value projection matrices, respectively. 
%
%
Here, we omit the batch dimension $B$, layer normalization~\cite{layernorm}, and GELU~\cite{hendrycks2016gaussian} for brevity.
%

Given the query, key, and value, we define slot attention operation as follows:


\begin{align} 
    \M &= \K\Q^{\top}/\sqrt{D_h}.
    \label{eqn:attn_m}
\end{align}


\noindent We normalize the attention map $\M \in \mathbb{R}^{NT \times K}$ along the \emph{slot}-axis
fostering competitive learning of action and scene representations in slots:
\begin{align}
    \A(n,k) &= \frac{\exp{(\M(n,k)})}{\sum_{i=1}^{K}\exp{(\M(n,i))}}.
    \label{eqn:attn_a}
\end{align}
Here, $n$ is the key index and $k$ is the slot index.
%
Then we normalize $\A$ along the key-axis using the $L^2$ norm function, yielding $\hat{\A}$.
Then we attend to input features with the attention map $\hat{\A}$ as follows:


\begin{equation}
    \Z = \hat{\A}^{\top} \V.
\end{equation}


%
\noindent For each iteration $m \in [1,M]$,  we update the slots as follows:


\begin{align}
    \S &= \mlp(\S + \Z) + \S + \Z.
    \label{eqn:layer}
\end{align}


\noindent The slots progressively encode disentangled action and scene information with iterative updates as shown in \figref{overview} (b).
To the best of our knowledge, this is the first work that successfully utilizes slot attention to obtain disentangled representations in video recognition.


\noindent\textbf{Slot Matching.}
Among $K$ learnable slots, DE selects an action and a scene slot by solving a $K$ to $2$ matching problem.
We use the cross-entropy between a true label and a prediction as the cost function for the matching.
We use a classification head $\psi$, shared across the action and scene tasks, for the prediction.
%
After computing a $K \times 2$ cost matrix, we solve the matching problem using the Hungarian algorithm~\cite{kuhn1955hungarian}.
Please see the supplementary material for details.


\noindent\textbf{Disentangling encoder loss.}
For action and scene slot learning, we define the disentangling encoder loss with a unified head for predicting a $N_A+N_S$ dimensional vector as follows:
\begin{align}    
\label{eq:L_D}
    L_{DE} &= -\sum_{c=1}^{N_A+N_S} [y^{a}_c\log(\psi(\S_{action})) + y^{s}_c \log{\psi(\S_{scene})}].
\end{align}

\noindent Here, $y^a$ denotes the ground-truth action label and $y^s$ represents the ground-truth scene label. $\S_{action}$ is the action slot, and $\S_{scene}$ is the scene slot.
In cases where the dataset does not provide ground truth scene labels, we obtain $y^s$ by running a frozen off-the-shelf scene recognition model. 
For more details, please see the supplementary material.

\noindent\textbf{Limitation.}
Due to the spurious correlation between actions and scenes in training data, \emph{naively} using the slot attention does not fully disentangle action and scene.
As shown in \figref{overview} (b), different slots represent distinct regions of a video \eg action (human), object, and scene regions. 
Since the training data has a spurious correlation, the disentangling loss \eqnref{L_D} could be decreased even if an action slot and scene slot are assigned as opposite slots.
The resulting model would learn entangled representations, leading to poor performance on test data with action-scene combinations different from the training data.
To address the issue, we design AMD as described in the next subsection.

\subsection{Action Mask Decoder}
\label{sec:aux}
We introduce the Action Mask Decoder (AMD) as shown in \figref{overview} (c). 
Given an action slot, AMD learns to predict an action mask.
For the AMD to predict a high-quality mask, the input slot should contain action information not scene or object information.
Therefore, employing the AMD prevents the DE from learning entangled action and scene representations.
Since slots are complementary to each other in slot attention, learning good action representation by AMD encourages the DE to learn good scene representation as well.

Inspired by reconstruction-based representation learning methods~\cite{he2022mae,videomae}, we employ a lightweight decoder $\phi$. 
%
The decoder takes the action slots $\S_{action}$, and the pseudo-human mask $\hat{\H}$ as input and reconstructs an action mask.
To obtain pseudo-human masks, we can employ any off-the-shelf method~\cite{xie2021segformer,croitoru2017unsupervised,xie2019object,fame}.
We choose a simple mask extraction approach~\cite{fame} without learning.
We define the action mask decoding loss as follows:
\begin{align}    
    L_{AMD} = -\tilde{\H}\log(\phi(\S_{action})) - (1-\tilde{\H})\log(1 - \phi(\S_{action})).
    \label{eqn:l_mp}
\end{align} 
$\tilde{\H}\in\mathbb{R}^N$ is the temporally averaged version of pseudo-human masks $\hat{\H} \in \mathbb{R}^{NT}$.


\subsection{Training and Inference} 
\label{sec:tr_inf}

\noindent\textbf{Training.}
For model training, we define a total loss function as follows:
\begin{equation}
    L = L_{DE} + \alpha L_{AMD} + \beta L_{AG} + \gamma L_{cos}.
    \label{eqn:l_tot}
\end{equation}
Here, $L_{AG}$ is the attention guidance loss to further guide the action slot learning.
%
%
We define $L_{AG}$ as a $L^2$ loss between the attention map and the pseudo-human masks: $L_{AG} = ||\A(:,k_{action})-\hat{\H}||_2^2$, where $k_{action}$ is the action slot index.
%
We also incorporate the cosine similarity loss between every pair of slots to diversify slots: $L_{cos} =  \frac{1}{K} \sum_{i=1}^{K}\sum_{j\neq i} {[1-\cos(\S_{i},\S_{j})]}$. $\alpha$, $\beta$, and $\gamma$ are hyperparameters.

\noindent\textbf{Inference.}
During the inference, we use the disentangling encoder only.
We feed a video into the model to extract $K$ slots. 
Among the $K$ slots, we assign action and scene slots based on the highest output probability.
The linear classifier $\psi$ takes the action and scene slots to predict the action and scene labels.

%% file: figure/fig_overview.tex
\begin{figure*}[t]
\centering
    \includegraphics[width=\linewidth]{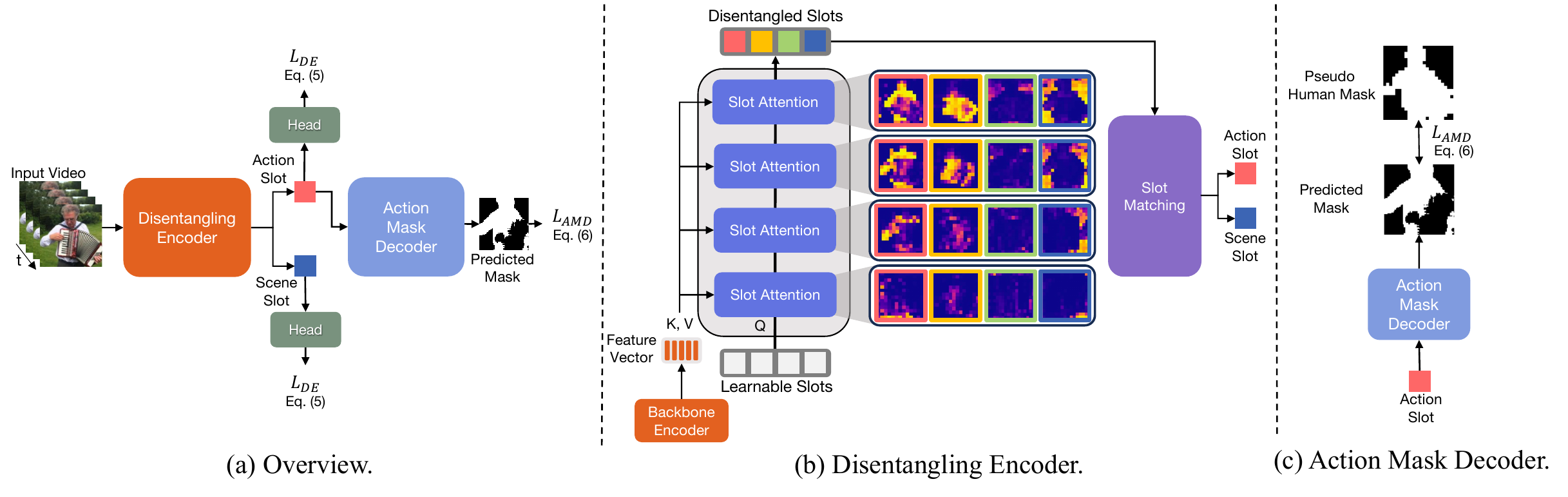}

\centering
\caption{
    \textbf{Overview of DEVIAS.} 
    (a)
    DEVIAS consists of i) a disentangling encoder (DE), ii) an action mask decoder (AMD), and iii) a classification head.
    (b) Given an input video, the DE first extracts a feature vector using a backbone encoder.
    %
    Then the DE learns multiple slots.
    %
    Given input learnable slots as queries, the slot attention iteratively attends to encoded features as keys and values.
    As a result of the slot attention, the slots progressively learn distinct information, \ie action, and scene.
    Then a matching function assigns each slot an action or a scene slot.
    We train action/scene slots with corresponding labels.
    (c) Given an action slot, the AMD predicts action masks to learn disentangled representations
    %
    As slots are complementary in slot attention, the AMD encourages the DE to learn not only good action but also good scene representations.
}
\label{fig:overview}
\end{figure*}



%% file: 5_result.tex
\section{Experimental Results}
\label{sec:results}
In this section, we carefully design and conduct extensive experiments to answer the following research questions: 
(1) Are the learned representations disentangled? (\secref{main_table}) 
%
%
(2) Are the disentangled representations beneficial for achieving a balanced action and scene recognition performance in seen and unseen action-scene combination scenarios?  (\secref{main_table}) 
%
%
(3) Are disentangled representations beneficial for transfer learning? (\secref{downstream})
(4) How can we disentangle representations? (\secref{abl}) 
To this end, we first provide details about the implementations in \secref{imp}, the datasets in \secref{dataset}, the evaluation metrics in \secref{eval}, and the baselines in \secref{baseline}.

\subsection{Implementation Details}
\label{sec:imp}
In this section, we briefly explain the implementation details. For the comprehensive details, please refer to the supplementary material.


\noindent\textbf{Training.} 
We densely sample 16 frames from each video to construct an input clip. 
We apply random cropping and resizing to every frame to get $224\times224$ pixels for each frame. 
We employ the ViT~\cite{vit} pre-trained with self-supervised VideoMAE~\cite{videomae} on the target dataset, \eg UCF-101 in \tabref{ucf_main}, and Kinetics-400 in \tabref{k400_main}, as our backbone encoder. 
We employ a 3-layer MLP as our action mask decoder.
We employ an off-the-shelf and non-learning-based pseudo-human mask extractor~\cite{fame} which computes foreground probability using pixel value statistics.
In cases where the ground-truth scene labels are not provided by the dataset, \eg UCF-101 and Kinetics-400, we employ the ViT pre-trained on the Places365~\cite{places365} dataset as the frozen scene model to obtain pseudo scene labels.


\noindent\textbf{Inference.} 
During inference, we average predictions over multiple temporal views and spatial crops, resulting in 2$\times$3 views in all experiments.


\subsection{Datasets}
\label{sec:dataset}

In this section, we briefly describe the datasets used. 
For the complete details, please refer to the supplementary material.
We evaluate DEVIAS across both \emph{seen} and \emph{unseen} action-scene combination scenarios, employing the standard training splits of either UCF-101~\cite{soomro2012ucf101} or Kinetics-400~\cite{kay2017kinetics} for model training.

\noindent\textbf{Seen combination.}
We use the original validation split of either UCF-101~\cite{soomro2012ucf101} or Kinetics-400~\cite{kay2017kinetics} for testing. 

\noindent\textbf{Unseen combination.}
%
We use the two synthetic datasets, SCUBA and HAT, to test the models in diverse unseen action-scene combination scenarios.
SCUBA~\cite{li2023stillmix} and HAT~\cite{chung2022hatdataset} contain diverse combinations in several dataset versions, denoted as SCUBA VQGAN-CLIP/Sinusoidal and HAT Random/Far.


\input{figure/fig_dataset}

\noindent\textbf{Realistic dataset.}
To evaluate performance using more realistic data, we rearrange the holistic video understanding (HVU) dataset~\cite{diba2020hvu} which provides both action and scene labels.
We train models on the train split of the HVU.
We test the models on the seen and unseen combination splits we rearranged.
Specifically, from the HVU validation set, we select videos with the same action-scene combinations as the training set to construct a seen combination split.
We select videos with the different action-scene combinations from the training set to construct an unseen split.
We show example frames of the datasets in the \figref{dataset}.

\noindent\textbf{Downstream datasets.}
We pre-train models on the Kinetics-400.
For fine-tuning, we use both temporal-biased datasets, the Something-Something V2~\cite{goyal2017something}, and Diving48~\cite{li2018resound}, and scene-biased datasets, UCF-101~\cite{soomro2012ucf101}, and ActivityNet~\cite{caba2015activitynet}.
%

\subsection{Evaluation Metric}
\label{sec:eval}

We report action and scene recognition performance across both \emph{seen} and \emph{unseen} action-scene combination scenarios.
We use top-1 action recognition accuracy for all the datasets and top-1 scene recognition for the HVU dataset.
%
%
Since UCF-101 and Kinetics-400 datasets do not have scene labels, we resort to using pseudo-labels generated by a Places365~\cite{places365} pre-trained scene model.
Given the fine-grained nature of the Places365 categories, we report top-5 accuracy for UCF-101 and Kinetics-400 scene recognition. 
To gauge a model's balanced performance in both seen and unseen combinations scenarios, we report the \emph{harmonic mean} (H.M.) across four performance metrics as our main metric: i) seen, and ii) unseen combinations action, iii) seen, and iv) unseen combinations scene.

\subsection{Baselines} 
\label{sec:baseline}

\noindent\textbf{Multi-task supervision.}
For a \emph{fair} comparison, we mainly compare DEVIAS to the baselines with both action and scene supervision \emph{exactly same} as ours. 
%
The Two-Token baseline involves a feature encoder and two distinct learnable tokens: one for action and another for scene, appended to the input patches. 
%
%
%
%
We supervise the action and scene tokens using their respective ground-truth labels.
%
Additionally, we explore two variations of this baseline, each incorporating either the BE~\cite{BE} or FAME~\cite{fame} debiasing methods.
%
%
Unlike the Two-Token approach, the One-Token baseline has a single token and employs a multi-task loss combining both action and scene losses to learn the token. 
We provide detailed descriptions and accompanying figures of the baselines in the supplementary materials.

\noindent\textbf{Single-task supervision.}
We compare DEVIAS with the baselines with single-task supervision for reference.
There are two naive ViT baselines trained using only action or scene labels and two scene-debiasing baselines: BE~\cite{BE} and FAME~\cite{fame} trained only with action labels. 
%

\subsection{Quantitative Analysis}
\label{sec:main_table}

\input{table/knn_reverse}

\noindent\textbf{Sanity-check: k-NN experiments.}
Before the main experiments, we check the sanity of DEVIAS.
For representation learning, we train DEVIAS and the baselines on the UCF-101~\cite{soomro2012ucf101} train split.
Then we store the action and scene feature vectors from all the training videos in the target datasets, UCF-101 and HMDB-51~\cite{kuehne2011hmdb}.
For k-NN testing, we evaluate performance in two scenarios: using the same feature types as in training (k-NN Normal Features), and using the alternate feature types (k-NN Reverse Features), as shown in \tabref{knn_reverse}.

\eg `A-S' indicates training the k-NN classifier with action features and testing with scene features.
We anticipate that if a model has disentangled action and scene representations, it would perform well in the k-NN Normal Features scenario, and show near-random performance in the k-NN Reverse Features scenario.
In \tabref{knn_reverse}, we compare our method with the One-Token and Two-Token baselines and the random performance. 

For the experiments, we employ a 10-NN setting and report top-1 accuracy.
We train both the baseline and DEVIAS on UCF-101.
The upper bound represents the performance achieved with supervised training on each dataset. 
In the k-NN Normal Features scenario, all the methods compared show reasonable action and scene recognition performances.
However, in the k-NN Reverse scenario, only DEVIAS shows near-random performance.
The results verify the disentangled action and scene representations of DEVIAS.

%
%

\noindent\textbf{UCF-101 results.}
In \tabref{ucf_main}, we evaluate the methods on the UCF-101 dataset.
%
%
Notably, the action performance of the Two-Token baseline decreases compared to the Naive Action ViT baseline in unseen combination scenarios: $16.7\%$ \vs $15.5\%$. 
%
The results indicate that the Two-Token baseline still learns scene-biased representations.
When we apply the scene debiasing techniques to the Two-Token baseline, we observe action performance improvement of $1.3 \sim 4.8$ points in unseen combination scenarios, and $2.4 \sim 7.0$ points improvement in H.M. compared to the Two-Token without debiasing.
%
%
DEVIAS stands out by showing a more balanced performance of action and scene, achieving a significant $16.4$ points boost over the second-best method in the H.M.. 
%
%
Remarkably, as a single model, DEVIAS surpasses the oracle performance of individual Action ViT with debiasing (BE) and Naive Scene ViT models: 50.3\% \vs 61.1\%.

\input{table/ucf101}
\input{table/kinetics400}

\noindent\textbf{Kinetics-400 results.}
In \tabref{k400_main}, we present the experimental results on the Kinetics-400 dataset, where DEVIAS shows a $1.6$ points improvement in H.M. over the second-best method. 
The overall trend among methods remains similar to the trend we observe in the UCF-101 dataset.

\noindent\textbf{HVU results.}
%
%
%
To further validate DEVIAS using \emph{more realistic data}, we show the results on the HVU dataset~\cite{diba2020hvu} in \tabref{hvu}. 
DEVIAS achieves the best performance in both action and scene recognition performance compared to baselines, showing a $2.4$ points improvement in H.M. over the second-best method.
The results demonstrate the effectiveness of DEVIAS when using more realistic data.

\input{table/rebuttal/hvu}

\input{table/downstream}


\subsection{Downstream Task}
\label{sec:downstream}

We investigate whether the disentangled action and scene representation is beneficial or not in various downstream tasks.
%
We conduct a set of transfer learning experiments: using the model weights trained on the source dataset as the initialization, we fine-tune the models on the target datasets.
For the target dataset fine-tuning, we use only the cross-entropy loss with action ground-truth labels.


We compare DEVIAS with naive baselines (Action ViT, Scene ViT), scene debiasing methods (BE~\cite{BE}, FAME~\cite{fame}), and the multi-task baselines (Two-Token, Two-Token w/ FAME).
%
%
For fine-tuning DEVIAS, we concatenate the action and scene tokens and feed the feature vector into the classification head.
%
%
In \tabref{downstream}, DEVIAS shows favorable performance on the downstream tasks across the temporal-biased and scene-biased tasks compared to the baselines.
%
%

\subsection{Ablation Study}
\label{sec:abl}


We conduct extensive ablation studies to validate the efficacy of each proposed module and design choices. 
We train all models on the UCF-101~\cite{soomro2012ucf101} train split. 
For evaluating action performance, we report top-1 accuracy on the validation split of UCF-101 (seen), and SCUBA-VQGAN-CLIP \cite{li2023stillmix} (unseen).
%
For scene performance, we report top-5 accuracy on the validation split of UCF-101 (seen) and UCF-101-Scene-only~\cite{chung2022hatdataset} (unseen). 
We report the harmonic mean of the four performances to assess the balanced performance of action and scene recognition.
Please refer to the supplementary material for more ablation experiments.

\input{table/ablation}
\noindent\textbf{Effect of the disentangling encoder.}
We investigate the effect of the DE. 
%
%
As shown in \tabref{abl} (a), incorporating the DE results in $13.4$ points enhancement in the H.M. compared to the baseline without the DE (with the AMD).


\noindent\textbf{Effect of the action mask decoder.}
%
In \tabref{abl} (b), we investigate the effect of the AMD.
Compared to the baseline without the AMD (with the DE), employing AMD shows a significant improvement of $5.9$ points in the H.M..
The results in \tabref{abl} (a) and (b), underscore the importance of employing \emph{both DE and AMD} for effective disentanglement of action and scene representations.


\noindent\textbf{Ablation study on the decoder design choices.}
In \tabref{abl} (c), we investigate various decoder design choices. 
%
In the first column, `Pixel' and `Mask' denote an AMD reconstructing RGB pixel values and action masks, respectively.
%
%
Compared to the action pixel reconstruction, the action mask reconstruction shows $2.6$ points improvement. 
Decoding the action slot only shows the best performance ($60.7\%$) compared to decoding the scene slot only ($53.1\%$) and decoding both the action and scene slots ($52.8\%$).



\noindent\textbf{Effect of hyperparameters.}
%
Increasing the number of slot attention iterations improves the H.M. by $4.9$ points as shown in the first and second rows in \tabref{abl} (d).
Using shared parameters for slot attention shows $5.6$ points improvement compared to using separate parameters for each slot attention layer.
We see a decrease in performance when using more slots, as shown in the fourth row.



\noindent\textbf{Effect of slot assignment method.} 
We examine the Hungarian matching when assigning disentangled slots as action or scene slots in \tabref{abl} (g). 
%
%
We observe a superior performance of 60.7 points H.M. when training with Hungarian matching, compared to both hard assignment and greedy matching of each slot to action and scene roles.
%


\noindent\textbf{Effect of softmax normalization axis.}
In \tabref{abl} (h), we analyze the effect of the softmax normalization axis.
Applying softmax normalization along the slot-axis, as opposed to the conventional key-axis normalization, results in a gain of $4.5$ points in the H.M.
The results indicate that slot attention, by competitively isolating features, significantly contributes to disentanglement.



\noindent\textbf{Effect of mask extraction method.}
Throughout our experiments, we utilize a simple mask extraction approach without learning~\cite{fame} by default. 
However, when using a learned segmentation method \eg SegFormer~\cite{xie2021segformer}, we observe a further improvement: $3.0$ points increase in H.M. as shown in \tabref{abl} (i).


%% file: figure/fig_dataset.tex

\setlength{\columnsep}{10pt}%
\setlength{\intextsep}{10pt}%
\begin{figure*}[h]
    \mpage{0.15}{
        \centering
        \includegraphics[width=\linewidth]{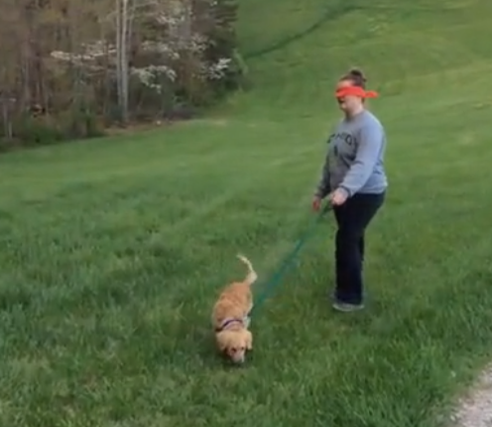}
    }
    \hfill
    \mpage{0.15}{
        \centering
        \includegraphics[width=\linewidth]{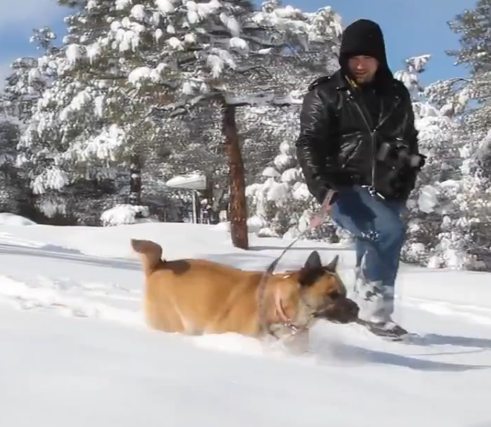}
    }  
    \hfill
    \mpage{0.15}{
        \centering
        \includegraphics[width=\linewidth]{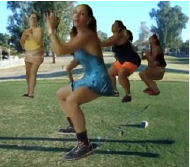}
    }    
    \hfill
    \mpage{0.15}{
        \centering
        \includegraphics[width=\linewidth]{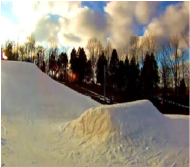}
    }
    \hfill
    \mpage{0.15}{
        \centering    
        \includegraphics[width=\linewidth]{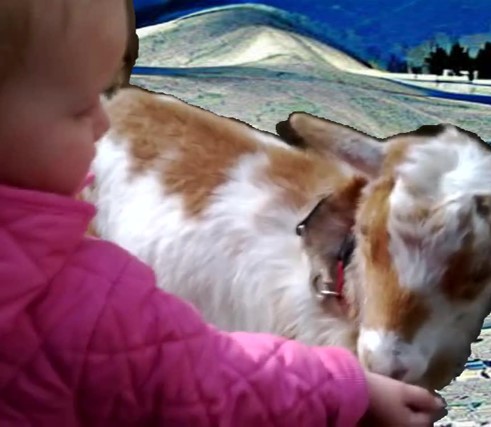}
    }  
    
    \figcapmargin
    \mpage{0.15}{
        {\fontsize{7pt}{10pt}\selectfont (a)}
    }
    \hfill
    \mpage{0.15}{
        {\fontsize{7pt}{10pt}\selectfont (b)}
    }    
    \hfill
    \mpage{0.15}{
        {\fontsize{7pt}{10pt}\selectfont (c)}
    }
    \hfill
    \mpage{0.15}{
        {\fontsize{7pt}{10pt}\selectfont (d)}
    }
    \hfill
    \mpage{0.15}{
        {\fontsize{7pt}{10pt}\selectfont (e)}
    }

    \captionof{figure}{\textbf{Example frames of the datasets.} (a) \emph{walking with dog} on the \emph{grass} from HVU Seen~\cite{diba2020hvu}, (b) \emph{walking with dog} in \emph{snowfield} from HVU Unseen ~\cite{diba2020hvu}, (c) \emph{dancing} on a \emph{golf course} from HAT Far~\cite{chung2022hatdataset}, (d) \emph{snowfield} from HAT Scene-Only~\cite{chung2022hatdataset}, and (e) \emph{feeding goats} from SCUBA VQGAN-CLIP~\cite{li2023stillmix}. 
    }

    \label{fig:dataset}

\end{figure*}

%% file: table/knn_reverse.tex
\begin{table}[t]

\centering
\caption{\textbf{Sanity check of disentanglement using k-NN.}  We compare models trained on the UCF-101~\cite{soomro2012ucf101} using k-NN classification accuracy (\%) on the UCF-101 and HMDB-51~\cite{kuehne2011hmdb}. For each column, we show \{\texttt{train}\}-\{\texttt{test}\} features of k-NN. `A' and `S' represent action and scene, respectively. The \best{best} performance is highlighted.}

\label{tab:knn_reverse}
\resizebox{0.8\linewidth}{!}{
\setlength{\tabcolsep}{8pt}
\begin{tabular}{ l c c c c c c c c c c c }
      \toprule
\multirow{3}{*}{Method} & \multicolumn{5}{c}{k-NN Normal Features (↑)} && \multicolumn{5}{c}{k-NN Reverse Features (↓)} \\
\cmidrule{2-6} 
\cmidrule{8-12}
& \multicolumn{2}{c}{UCF-101} && \multicolumn{2}{c}{HMDB-51} && \multicolumn{2}{c}{UCF-101} && \multicolumn{2}{c}{HMDB-51} \\
\cmidrule{2-3} \cmidrule{5-6}  
\cmidrule{8-9} \cmidrule{11-12}
& A-A & S-S && A-A & S-S && A-S & S-A && A-S & S-A \\
\midrule
Random  &1.0 & 0.3 && 1.0 & 0.3 && 1.0 & 0.3 && 1.0 & 0.3 \\
\midrule
One-Token & 87.7 &43.5 && \cellcolor{gray!30}\best{43.4} & \cellcolor{gray!30}\best{27.5} && 87.7 &43.5 && 43.4 & 27.5  \\
Two-Token &   84.2 & \cellcolor{gray!30}\best{43.6} && 37.4 &25.7 &&62.8 & 37.1 && 27.3 &20.7\\
DEVIAS & \cellcolor{gray!30}\best{89.7} & 41.8 && 38.8 & 26.3 && \cellcolor{gray!30}\best{4.5} & \cellcolor{gray!30}\best{0.3} && \cellcolor{gray!30}\best{1.9} & \cellcolor{gray!30}\best{0.9} \\

\midrule
Upperbound & 92.0 & 48.8 && 60.5 & 38.1 && - & - && - & - \\
\bottomrule
\end{tabular}
}

\end{table}




%% file: table/ucf101.tex
\begin{table*}[t]
\centering
\caption{\textbf{Action and scene recognition performance on the UCF-101 dataset}. We report the Top-1 action recognition and the Top-5 scene recognition accuracies (\%). We evaluate both seen and unseen action-scene combination scenarios. We also report the harmonic mean (H.M.) of the action recognition and scene recognition. V.C./Sin. denotes the SCUBA~\cite{li2023stillmix} VQGAN-CLIP/Sinusoidal; S.O./Rand. denotes the HAT~\cite{chung2022hatdataset} Scene-Only/Random. The \best{best} and the  \second{second-best} H.M. numbers are highlighted.
}

\def\arraystretch{1.2}
\resizebox{1.0\linewidth}{!}{
\setlength{\tabcolsep}{3pt}
\begin{tabular}{ lccl cccc  c ccccc c}
\toprule
 
\multirowcell{3}[-1.5ex]{\shortstack{Training \\ Strategy}} & \multicolumn{2}{c}{\multirowcell{2}[-0.7ex]{Supervision}} &  & \multicolumn{4}{c}{Action($\uparrow$) } && \multicolumn{5}{c}{Scene($\uparrow$)} & \\
\cmidrule{5-8}
\cmidrule{10-14}
& && \multicolumn{1}{c}{Method} & \multirow{2}{*}{\shortstack{Seen}} & \multicolumn{3}{c}{Unseen} && \multirow{2}{*}{\shortstack{Seen}} & \multicolumn{4}{c}{Unseen} & H.M. \\
\cmidrule{2-3} \cmidrule{6-8} \cmidrule{11-14} 
& Action&Scene&&   & V.C. & \multicolumn{1}{c}{Sin.} & Mean &&  &S.O.& Rand. & Far & Mean & \\
\midrule

\multirow{4}{*}{Single-Task} &\checkmark&$\times$  &  Naive Action ViT  &  \cellcolor{gray!30}92.9  &  12.4 &  \multicolumn{1}{c}{21.0} &\cellcolor{gray!30}16.7 &&\cellcolor{gray!30}-&-&-&-&\cellcolor{gray!30}-&\cellcolor{gray!30}-\\
& \checkmark &$\times$& BE \cite{BE}  &  \cellcolor{gray!30}92.3   & 12.1  & \multicolumn{1}{c}{38.7} &\cellcolor{gray!30}25.4 &&\cellcolor{gray!30}-&-&-&-&\cellcolor{gray!30}-&\cellcolor{gray!30}-\\
 &\checkmark&$\times$ & FAME \cite{fame} &  \cellcolor{gray!30}91.6  & 24.8  &  \multicolumn{1}{c}{15.6}  &\cellcolor{gray!30}20.2 &&\cellcolor{gray!30}-&-&-&-&\cellcolor{gray!30}-&\cellcolor{gray!30}-\\


&$\times$ &\checkmark & Naive Scene ViT &\cellcolor{gray!30} -&-&-&\cellcolor{gray!30}-&&  \cellcolor{gray!30}72.0 & 61.7 & 62.8 &\multicolumn{1}{c}{69.6} & \cellcolor{gray!30}64.7&\cellcolor{gray!30}-\\

\midrule

 \multirow{4}{*}{Multi-Task}&\checkmark&\checkmark &  One-Token &  \cellcolor{gray!30}91.9 & 5.0 & \multicolumn{1}{c}{21.8} & \cellcolor{gray!30}13.4 & &  \cellcolor{gray!30}74.0 & \multicolumn{1}{c}{60.5} & 58.0 & 66.5 & \cellcolor{gray!30}61.7 & \cellcolor{gray!30}34.7 \\
 &  \checkmark &\checkmark & Two-Token &  \cellcolor{gray!30}86.0  &11.1     & \multicolumn{1}{c}{19.9} &\cellcolor{gray!30}15.5 &&  \cellcolor{gray!30}72.3 & 59.6 & 59.2 &\multicolumn{1}{c}{67.1} &\cellcolor{gray!30}62.0&  \cellcolor{gray!30}37.7\\
  &\checkmark &\checkmark & Two-Token w/ BE \cite{BE} & \cellcolor{gray!30}89.9   &  13.0   & \multicolumn{1}{c}{20.5} &\cellcolor{gray!30}16.8 &&  \cellcolor{gray!30}74.2  & 62.3 & 59.3 & \multicolumn{1}{c}{69.5} & \cellcolor{gray!30}63.7& \cellcolor{gray!30}40.1\\
 & \checkmark &\checkmark& Two-Token w/ FAME \cite{fame} &  \cellcolor{gray!30}89.5   &25.3    &\multicolumn{1}{c}{15.3} &\cellcolor{gray!30}20.3  &&  \cellcolor{gray!30}73.2 & 61.4 & 62.8 & \multicolumn{1}{c}{70.3} & \cellcolor{gray!30}64.8& \cellcolor{gray!30}\second{44.7}\\
\midrule
 Disentangle&\checkmark&\checkmark & DEVIAS & \cellcolor{gray!30}90.1   &40.1    &  \multicolumn{1}{c}{38.6} &\cellcolor{gray!30}39.4 &&  \cellcolor{gray!30}74.0  &  61.0 & 62.4 & \multicolumn{1}{c}{70.2} &\cellcolor{gray!30}64.5 & \cellcolor{gray!30}\best{61.1}\\
\bottomrule
\end{tabular}
}
\label{tab:ucf_main}
\end{table*}

%% file: table/kinetics400.tex
\begin{table}[t]
\centering
\caption{\textbf{Action and scene recognition performance on the Kinetics-400}. We report the Top-1 action recognition and the Top-5 scene recognition accuracies (\%). We evaluate both seen and unseen action-scene combination scenarios. We also report the harmonic mean (H.M.) of the action recognition and scene recognition. V.C./Sin. denotes the SCUBA~\cite{li2023stillmix} VQGAN-CLIP/Sinusoidal; S.O./Rand. denotes the HAT~\cite{chung2022hatdataset} Scene-Only/Random. The  \best{best} and the  \second{second-best} H.M. numbers are highlighted. 
}

\def\arraystretch{1.2}

\resizebox{1.0\linewidth}{!}{
\setlength{\tabcolsep}{3pt}
\begin{tabular}{ lccl cccc  c ccccc c}
\toprule
 
\multirowcell{3}[-1.5ex]{\shortstack{Training \\ Strategy}} & \multicolumn{2}{c}{\multirowcell{2}[-0.7ex]{Supervision}} &  & \multicolumn{4}{c}{Action($\uparrow$) } && \multicolumn{5}{c}{Scene($\uparrow$)} & \\
\cmidrule{5-8}
\cmidrule{10-14}
& && \multicolumn{1}{c}{Method} & \multirow{2}{*}{\shortstack{Seen}} & \multicolumn{3}{c}{Unseen} && \multirow{2}{*}{\shortstack{Seen}} & \multicolumn{4}{c}{Unseen} & H.M. \\
\cmidrule{2-3} \cmidrule{6-8} \cmidrule{11-14} 
& Action&Scene&&   & V.C. & \multicolumn{1}{c}{Sin.} & Mean &&  &S.O.& Rand. & Far & Mean & \\
\midrule

\multirow{4}{*}{Single-Task} &\checkmark&$\times$ & Naive Action ViT  &  \cellcolor{gray!30}76.8  &  41.6 &  \multicolumn{1}{c}{49.6} &\cellcolor{gray!30}45.6 &&\cellcolor{gray!30}-&-&-&-&\cellcolor{gray!30}-&\cellcolor{gray!30}-\\
&\checkmark&$\times$& BE \cite{BE}  &  \cellcolor{gray!30}77.6&        43.2& \multicolumn{1}{c}{52.2} &\cellcolor{gray!30}47.7&&\cellcolor{gray!30}-&-&-&-&\cellcolor{gray!30}-&\cellcolor{gray!30}-\\
 &\checkmark&$\times$& FAME \cite{fame} & \cellcolor{gray!30}77.8 & 49.7 &  \multicolumn{1}{c}{56.8}  &\cellcolor{gray!30}53.3 &&\cellcolor{gray!30}-&-&-&-&\cellcolor{gray!30}-&\cellcolor{gray!30}-\\


&$\times$&\checkmark& Naive Scene ViT & \cellcolor{gray!30}-&-&-&\cellcolor{gray!30}-& & \cellcolor{gray!30}86.5&  82.6& \multicolumn{1}{c}{79.9} & 81.2 & \cellcolor{gray!30}81.2 &\cellcolor{gray!30}-\\

\midrule

\multirow{4}{*}{Multi-Task}  &\checkmark&\checkmark&    One-Token & \cellcolor{gray!30}74.2  & 35.2 & \multicolumn{1}{c}{45.6} & \cellcolor{gray!30}40.4 & & \cellcolor{gray!30}87.9 & \multicolumn{1}{c}{83.8} & 80.8 & 81.5 & \cellcolor{gray!30}82.0 & \cellcolor{gray!30}64.7\\
& \checkmark&\checkmark&  Two-Token & \cellcolor{gray!30}75.0  & 34.9 & \multicolumn{1}{c}{46.6} &\cellcolor{gray!30}40.8 && \cellcolor{gray!30}86.4 & 75.8 & \multicolumn{1}{c}{78.3} & 80.3 &\cellcolor{gray!30}78.1&  \cellcolor{gray!30}64.3\\
   & \checkmark&\checkmark& Two-Token w/ BE \cite{BE} & \cellcolor{gray!30}75.1  & 35.8 & 48.0& \cellcolor{gray!30}41.9 & & \cellcolor{gray!30}87.7 & 82.9 & 80.0 & 81.5 &\cellcolor{gray!30}81.5 &\cellcolor{gray!30}65.7\\
 &\checkmark&\checkmark & Two-Token w/ FAME \cite{fame} & \cellcolor{gray!30}75.0 & 45.5 &\multicolumn{1}{c}{51.5} &\cellcolor{gray!30}48.5  && \cellcolor{gray!30}87.3 & 77.4 &\multicolumn{1}{c}{81.1} & 82.6 & \cellcolor{gray!30}80.4& \cellcolor{gray!30}\second{69.2}\\
 \midrule
 Disentangle& \checkmark&\checkmark& DEVIAS & \cellcolor{gray!30}77.3 & 50.3&  \multicolumn{1}{c}{58.8} &\cellcolor{gray!30}54.6&& \cellcolor{gray!30}82.0 &  76.5& \multicolumn{1}{c}{75.7} & 77.1 &\cellcolor{gray!30}76.4& \cellcolor{gray!30}\best{70.8}\\

\bottomrule
\end{tabular}

}
\label{tab:k400_main}
\end{table}

%% file: table/rebuttal/hvu.tex
\begin{table}[t]
  \centering
  \caption{\textbf{Action and scene recognition performance on the HVU dataset.} We report the Top-1 accuracy (\%) in both seen and unseen scenarios. We also report the harmonic mean (H.M.) of the action and scene performance. The \best{best} and the  \second{second-best} H.M. numbers are highlighted. 
  }
    
        \centering
        \resizebox{0.8\linewidth}{!}{
        \begin{tabular}{lccl cccc c}
            \toprule            
          \multirowcell{2}[-1ex]{\shortstack{Training \\ Strategy}}&\multicolumn{2}{c}{Supervision}& \multicolumn{1}{c}{\multirow{2}{*}{Method}}& \multicolumn{2}{c}{Action($\uparrow$)} & \multicolumn{2}{c}{Scene($\uparrow$)} & \multirow{2}{*}{H.M.}\\
         \cmidrule(lr){2-3} \cmidrule(lr){5-6} \cmidrule(lr){7-8} 

       &Action&Scene& & Seen & Unseen & Seen & Unseen \\

            \midrule
       \multirow{3}{*}{Single-Task} &\checkmark&$\times$&Naive Action ViT & 82.5 & 34.9 & -&-&\cellcolor{gray!30}-\\
        &\checkmark&$\times$& FAME~\cite{fame} & 81.0 & 35.4 &  -&-&\cellcolor{gray!30}-\\
        &$\times$ &\checkmark & Naive Scene ViT & - & - & 97.0 & 45.4 & \cellcolor{gray!30}- \\
        \midrule
       \multirow{2}{*}{Multi-Task} &\checkmark&\checkmark&Two-Token & 80.5 & 34.9 & 98.5 & 45.9 & \cellcolor{gray!30}\second{54.8} \\
        &\checkmark&\checkmark& Two-Token w/ FAME \cite{fame} & 81.5 & 34.1 & 98.5 & 47.2 &\cellcolor{gray!30}\second{54.8}\\
        \midrule
       Disentangle&\checkmark&\checkmark  &DEVIAS & 83.5 & 36.2 & 99.0 & 49.3 & \cellcolor{gray!30}\best{57.2}\\ 
        
           \bottomrule
        \end{tabular}
        }
\label{tab:hvu}
\end{table}

%% file: table/downstream.tex
\begin{table*}[t]
\centering
\caption{\tb{Downstream task performance.} We report Top-1 accuracy (\%). All models are pre-trained on the Kinetics-400 and then fine-tuned on the downstream datasets. SSV2 denotes the Something-Something-V2 dataset. The \best{best} and the \second{second-best} H.M. numbers are highlighted. 
}

\resizebox{0.8\linewidth}{!}{
\begin{tabular}{lll cc c cc c}
    
\toprule
          \multirowcell{2}{\shortstack{Pretraining \\ Strategy}} && \multicolumn{1}{c}{\multirow{2}{*}{Method}}&\multicolumn{2}{c}{Temporal-biased}&&\multicolumn{2}{c}{Scene-biased} &\\
\Xcline{4-5}{0.03em}
\Xcline{7-8}{0.03em}
 &&& Diving48 & SSV2&  & UCF-101 & ActivityNet&\multirow{-2}{*}{H.M.}\\
\midrule
\multirow{4}{*}{Single-Task} && Naive Action ViT  & 81.5 &74.2&& 98.5 & 84.4&\cellcolor{gray!30}83.8 \\
&&BE~\cite{BE} & 81.9 &74.5&& 98.3 & 84.6 & \cellcolor{gray!30}\second{84.0}\\
&&FAME~\cite{fame} & 80.6 &74.2& & 98.3 & 83.8& \cellcolor{gray!30}83.4\\
&&Naive Scene ViT  & 73.1 & 71.8 && 92.0 & 73.1 & \cellcolor{gray!30}76.7 \\
\midrule
\multirow{2}{*}{Multi-Task}&&Two-Token & 80.1 & 73.7&&98.2&83.7&\cellcolor{gray!30}83.0\\
&&Two-Token w/ FAME~\cite{fame} & 78.7 & 73.5 & &98.1 & 81.5 & \cellcolor{gray!30}82.0 \\
\midrule
Disentangle&&DEVIAS & 84.4 & 75.2 && 98.4 & 84.5 & \cellcolor{gray!30}\textbf{84.8} \\

\bottomrule
\end{tabular}
}
\label{tab:downstream}
\label{tab:action_noaug_hyper}
\end{table*}





%% file: table/ablation.tex

\begin{table*}[t]
  \centering
  \caption{\textbf{Ablation study.} To validate the effect of each component, we show the results on the UCF-101 dataset. In every experiment, we use a ViT backbone pre-trained on the UCF-101. We report the Top-1 accuracy (\%) for the action and the Top-5 accuracy (\%) for the scene recognition, along with the harmonic mean (H.M.) of the two accuracies. The \best{best} numbers are highlighted.}


    \mpage{0.48}{\scriptsize (a) Effect of disentangling encoder.}\hfill
    \mpage{0.48}{\scriptsize (b) Effect of action mask decoder.}

    \mpage{0.48}{
        \centering
        \resizebox{\linewidth}{!}{
        \begin{tabular}{lccccc}
            \toprule
            \multirow{2}{*}{Method}& \multicolumn{2}{c}{Action($\uparrow$)} & \multicolumn{2}{c}{Scene($\uparrow$)} & \multirow{2}{*}{H.M.}\\
            \cmidrule(lr){2-3} \cmidrule(lr){4-5} 
           &Seen&  Unseen & Seen & Unseen \\
            \midrule
           
           
           w/o disentangling encoder & 89.1 & 23.7 & 71.5 & 58.7 & \cellcolor{gray!30}47.3\\
           w/ disentangling encoder & 90.1 & 40.1 & 74.0 & 61.0 & \cellcolor{gray!30}\best{60.7} \\
           \bottomrule
        \end{tabular}
        }
    }
    \hfill
    \mpage{0.48}{
        \centering
        \resizebox{\linewidth}{!}{




    
        \begin{tabular}{lccccc}
            \toprule
            \multirow{2}{*}{Method}& \multicolumn{2}{c}{Action($\uparrow$)} & \multicolumn{2}{c}{Scene($\uparrow$)} & \multirow{2}{*}{H.M.}\\
            \cmidrule(lr){2-3} \cmidrule(lr){4-5} 
           &Seen&  Unseen & Seen & Unseen \\
            \midrule
           w/o action mask decoder & 90.0 & 31.6 & 73.7 & 59.8 & \cellcolor{gray!30}54.8 \\
           w/ action mask decoder & 90.1 & 40.1 & 74.0 & 61.0 & \cellcolor{gray!30}\best{60.7} \\
           \bottomrule
        \end{tabular}
        }
    }
                
               
                

    \mpage{0.48}{\scriptsize (c) Ablations on decoder design choices.}\hfill           
    \mpage{0.48}{\scriptsize (d) Effect of hyperparameters.}

    \mpage{0.48}{
        \centering
        \resizebox{0.95\linewidth}{!}{
        \begin{tabular}{cccccccc}
            \toprule
            \multirow{2}{*}{Target} & \multirow{2}{*}{Action} & \multirow{2}{*}{Scene}&  \multicolumn{2}{c}{Action($\uparrow$)}&\multicolumn{2}{c}{Scene($\uparrow$)}&\multirow{2}{*}{H.M.}\\
             \cmidrule(lr){4-5} \cmidrule(lr){6-7} 
            
            && & Seen & Unseen & Seen & Unseen \\
            \midrule
            Pixel&\checkmark&$\times$& 89.7 & 36.4 & 72.5 & 61.0 & \cellcolor{gray!30}58.1 \\
            Mask&\checkmark&$\times$&90.1 & 40.1 & 74.0 & 61.0 & \cellcolor{gray!30}\best{60.7}\\
            Mask&$\times$&\checkmark& 89.4 & 28.9 & 74.7 & 61.9 & \cellcolor{gray!30}53.1\\
            Mask&\checkmark&\checkmark& 89.3 & 28.6 & 74.6 & 61.6 & \cellcolor{gray!30}52.8 \\
            
            \bottomrule
        \end{tabular}
        }
    }
    \hfill    
    \mpage{0.48}{
        \centering
        \resizebox{\linewidth}{!}{
        \begin{tabular}{cccccccc}
            \toprule
            \multicolumn{3}{c}{Hyperparameters}&  \multicolumn{2}{c}{Action($\uparrow$)}&\multicolumn{2}{c}{Scene($\uparrow$)}&\multirow{2}{*}{H.M.}\\
            \cmidrule(lr){1-3} \cmidrule(lr){4-5} \cmidrule(lr){6-7} 
            
            No. Slots & No. Iter. & Shared? & Seen & Unseen & Seen & Unseen \\
            \midrule
           
             2&2&\checkmark& 90.5 & 33.1 & 73.5 & 59.4 & \cellcolor{gray!30}55.8 \\
            2&4&\checkmark& 90.1 & 40.1  & 74.0 & 61.0 & \cellcolor{gray!30}\best{60.7}\\
            2&4&$\times$& 89.1 & 33.4 & 70.7 &57.7 & \cellcolor{gray!30}55.1 \\
             4&4&\checkmark& 89.1 & 39.5 & 71.3 & 58.6 & \cellcolor{gray!30}59.1\\
            
            \bottomrule
        \end{tabular}
        }
    }

    \mpage{0.32}{\scriptsize (g) Effect of slot assignment.}\hfill
    \mpage{0.32}{\scriptsize (h) Effect of softmax  axis.}\hfill    
    \mpage{0.32}{\scriptsize (i) Effect of mask extraction.}    
    \vspace{-1.0em}
    
    \mpage{0.32}{
        \centering
        \resizebox{\linewidth}{!}{
            \begin{tabular}{lccccc}
                \toprule
                \multirow{2}{*}{Method} & \multicolumn{2}{c}{Action($\uparrow$)}  & \multicolumn{2}{c}{Scene($\uparrow$)} & \multirow{2}{*}{H.M.}  \\
            \cmidrule(lr){2-3} \cmidrule(lr){4-5} 
                
                & Seen & Unseen & Seen & \multicolumn{1}{c}{Unseen}\\
                \midrule    
                Hard assign & 89.4 & 35.1 & 71.1 & 58.3 & \cellcolor{gray!30}56.4  \\
                Greedy & 87.2 & 37.1 & 69.7 & 56.7 & \cellcolor{gray!30}56.8 \\

                Hungarian & 90.1 & 40.1 & 74.0 & 61.0 & \cellcolor{gray!30}\best{60.7}\\
                
                \bottomrule
            \end{tabular}
        }
    }
    \hfill
    \mpage{0.32}{
        \resizebox{0.85\linewidth}{!}{
        \begin{tabular}{lccccc}
            \toprule
            \multirow{2}{*}{Axis} &  \multicolumn{2}{c}{Action($\uparrow$)} & \multicolumn{2}{c}{Scene($\uparrow$)} & \multirow{2}{*}{H.M.}\\
            \cmidrule(lr){2-3} \cmidrule(lr){4-5} 
            
            & Seen & Unseen & Seen  & Unseen\\
            \midrule
        
        
            Keys & 89.6 & 33.0 & 73.9 & 61.6 & \cellcolor{gray!30}56.2 \\
            Slots & 90.1 & 40.1 & 74.0 & 61.0 & \cellcolor{gray!30}\best{60.7} \\
            
            \bottomrule
        \end{tabular}
        }
    }
    \hfill
    \mpage{0.30}{
        \centering
        \resizebox{\linewidth}{!}{
        \begin{tabular}{lccccc}
            \toprule
            \multirow{2}{*}{Method}& \multicolumn{2}{c}{Action($\uparrow$)} & \multicolumn{2}{c}{Scene($\uparrow$)} & \multirow{2}{*}{H.M.}\\
            \cmidrule(lr){2-3} \cmidrule(lr){4-5} 
            
           &  Seen & Unseen & Seen& Unseen \\
            \midrule
            FAME \cite{fame} & 90.1 & 40.1 & 74.0 & 61.0 & \cellcolor{gray!30}60.7 \\
           Segformer \cite{xie2021segformer} & 90.0 & 46.6 & 73.8 & 59.9 & \cellcolor{gray!30}\best{63.7} \\
           \bottomrule
        \end{tabular}
        }
    }

\label{tab:abl}
\end{table*}

%% file: 6_conclusion.tex
\section{Conclusions}
\label{sec:conclusions}





In this paper, we tackle the under-explored problem of disentangled action and scene representation learning for holistic video understanding. 
We propose DEVIAS, a novel method that employs a disentangling encoder with the slot attention mechanism and action mask decoder to effectively learn disentangled action and scene representations. 
Through rigorous experiments, we assess both the action and scene recognition performance of DEVIAS in seen and unseen action-scene combination scenarios.
DEVIAS shows favorable performance compared to the baselines.
Furthermore, we demonstrate that disentangled action and scene representations are beneficial for various downstream tasks.
The results showcase the effectiveness of the DEVIAS in learning disentangled action and scene representations as a single model. 
We believe our work provides interesting insights into the video understanding community and will inspire future advancements.

%% file: 99_supp_content.tex
\noindent In this supplementary material, we provide comprehensive method/implemen-tation/baseline/dataset details, and quantitative/qualitative results to complement the main paper. We organize the supplementary material as follows:
\renewcommand{\thesection}{\Alph{section}}
\setcounter{section}{0}
\begin{enumerate}[label=\Alph*.]
    \item Additional details about DEVIAS
    \item Complete implementation details
    \item Compared method details
    \item Dataset details
    \item Evaluation metric
    \item Additional results
\end{enumerate}

\section{Additional Details about DEVIAS}
\label{sec:devias_detail}
In this section, we provide details of DEVIAS. In our implementation, we employ ViT-Base~\cite{vit} as the encoder. 

\noindent\textbf{Data augmentation.} 
\label{sec:data_aug}
In DEVIAS, we augment input videos to diversify action-scene combinations. 
Previous works~\cite{actorcut,fame} have demonstrated that scene augmentation by mixing up using a human mask improves action recognition.
We mix a video $\V_i$ with another video $\V_j$ within a minibatch, by cut-and-paste operation as follows:
\begin{equation}
\label{eq:Data_augmentation}
\tilde{\V}_i = \V_i  \odot \Tilde{\H}_i + \V_j  \odot (1-\Tilde{\H}_i),
\end{equation}
where , $\Tilde{\H}_i$ is a pseudo-human mask of the video $\V_i$ extracted by an off-the-shelf method~\cite{fame}. This simple scenario augmentation can diversify action-scene combinations in a training dataset.

\noindent\textbf{Scene model.} 
In cases where the dataset does not provide ground truth scene labels, we obtain scene labels by running a frozen off-the-shelf scene recognition model. 
We opt to use a ViT-Base as a frozen scene model to generate pseudo scene labels for each video in the action recognition datasets we use.
We prepare the frozen scene model by training a ViT-Base on the Places365~\cite{places365} dataset.
Given that we are dealing with video data, we inflate the static images of the Places365 dataset to introduce a temporal dimension. 
%
We generate a pseudo scene label, denoted as $y^s$, by taking the frozen scene model prediction on a video $\V$. 
We also provide additional results using various scene models in \secref{scene_model_exp}.

\noindent\textbf{Unified classification head.} We employ a unified linear classification head, $\psi$, designed to output a vector of dimension $N_A+N_S$, for predicting both action with $N_A$ classes and scene with $N_S$ classes. 

\noindent\textbf{Slot matching.}
Following the previous works~\cite{slot_object_centric,detr}, we employ the Hungarian algorithm~\cite{kuhn1955hungarian} as our matching function to assign each learned slot as either an action or scene based on the matching cost. 
We use the cross-entropy between a true label (or a pseudo label) and a prediction as the cost function for matching. 
In order to align the dimensions of the (pseudo) ground-truth label with those of the unified head, we zero-pad the end of the (pseudo) ground-truth label, extending it to match the combined length of $N_A + N_S$.

%
%
%
%
%

\begin{algorithm}
\caption{Slot attention}
\begin{algorithmic}[1] 
\label{slot_algo}
\STATE {\footnotesize \textbf{INPUT:} Slots $\S \in \mathbb{R}^{K \times D}$, Features $\X \in \mathbb{R}^{NT \times D}$, Number of iterations $M$, \\ and Hidden dimension $D_h$}
\STATE {\footnotesize \textbf{Layer params:} $\W_Q$, $\W_K$, $\W_V$: $\mathbb{R}^{D \times D_h}$ linear projections for attention; LayerNorm; $\text{Linear}$, GELU for MLP}
\FOR{$m = 1$ to $M$}
    \STATE $\Q = \W_Q(\text{LayerNorm}(\S))$ 
    \STATE $\K = \W_K(\text{LayerNorm}(\X))$ 
    \STATE $\V = \W_V(\text{LayerNorm}(\X))$ 
    \STATE $\A = \text{Softmax}\left(\frac{\K\Q^{\top}}{\sqrt{D_h}}, 
 \text{axis=`slots'}\right)$
    \STATE $\hat{\A}$ = L2Norm($\A, \text{axis=`key'}$)
    \STATE $\Z = \hat{\A}^{\top}\V$
    \STATE $\Z_{\text{mlp}} = \text{Linear}(\text{GELU}(\text{Linear}(\S+\Z)))$
    \STATE $\S = \Z_{\text{mlp}} + \S + \Z$
    
\ENDFOR
\RETURN $\S$
\end{algorithmic}
\end{algorithm}
\noindent\textbf{Slot attention.}
We provide a detailed pseudo-code for the slot attention in \algref{slot_algo}. 
Note that within the slot attention, all linear layers share weights.
%

\section{Complete Implementation Details}
In this section, we provide comprehensive details of our experimental setup and implementation details. We conduct the experiments with 24 NVIDIA GeForce RTX 3090 GPUs. We build upon the codebase of VideoMAE\footnote{\url{https://github.com/MCG-NJU/VideoMAE}}~\cite{videomae}. We implement DEVIAS using the DeepSpeed\footnote{\url{https://github.com/microsoft/DeepSpeed}} framework for faster training.

\noindent\textbf{Data preprocessing.} We densely sample frames from a video to obtain a clip of 3 channels $\times$ 16 frames $\times$ 224 width $\times$ 224 height. 
We set the frame interval as 4 frames for the dense sampling.
Given the sampled clip, we apply the data augmentation described in \secref{data_aug}.  
We use the augmented clip as an input to the encoder and the scene model.
Following VideoMAE~\cite{videomae}, we employ 3D convolution for patch embedding to effectively reduce the temporal dimension by half. 
This process results in a total of 8 $\times$ 196 tokens.
We maintain this data preprocessing protocol consistent across all experiments.

\noindent\textbf{Scene model training.} 
We employ a ViT-Base, pre-trained on the ImageNet-1k~\cite{imagenet} with the MAE~\cite{he2022mae} training method, fine-tuned on the Places365~\cite{places365} as the scene model. 
Given that we are dealing with video data, we inflate the static images of the Places365 dataset to introduce a temporal dimension.
%
%
We summarize the hyperparameters used in \tabref{scene_hyper}.

\noindent\textbf{Model training.}
We employ ViT-Base as our encoder. We use the ViT-Base pre-trained with the VideoMAE~\cite{videomae} training strategy on the dataset we use in the main paper, \eg UCF-101 and Kinetics-400. 
For the HVU~\cite{diba2020hvu} experiments, we use the ViT-Base pre-trained on the Kinetics-400 using the VideoMAE training strategy.
In the case of the UCF-101, we further fine-tune the pre-trained ViT-Base with action supervision, before the disentangled representation learning as we empirically find it is beneficial for stabilizing the training process.
We obtain the pseudo-human mask for UCF-101, Kinetics-400, and HVU datasets using FAME~\cite{fame} as it does not require any external object detector. 
FAME extracts the moving foreground region from the background regions using frame difference and color statistics.
In FAME, the parameter $\tau$ represents the threshold for the foreground ratio in the mask, selecting the top-k percentage as the foreground. 
For instance, a $\tau$ value of 0.4 indicates that the top 40\% of the mask is considered as foreground regions.
The augmentation ratio within a batch, denoted by $\rho$, determines the proportion of scene augmented data in a batch using \eqnref{Data_augmentation}.
For UCF-101, we set $\tau$ and $\rho$ to 0.3 and 0.4, respectively.
%
For Kinetics-400, we set them to 0.5 and 0.8. For HVU, we set them to 0.5 and 0.25. 
We linearly scale the base learning rate, then $\text{\emph{actual lr}} = \text{\emph{base lr}} \times  \text{\emph{Per GPU batch size}} \times  \text{\emph{number of GPUs}} /256 $.
We summarize the hyperparameters used in \tabref{hyper}.


\noindent\textbf{Inference.}
During inference, we sample an input video multiple times to generate multiple temporal views.
We resize each frame of each temporal view to $256 \times 256$. 
Subsequently, we crop the video multiple times to generate multiple spatial crops. 
The final prediction is obtained by aggregating predictions from (temporal views) $\times$ (spatial crops). 
For the UCF-101, Kinetics-400, and HVU datasets, we employ the (2 clips) $\times$ (3 crops) configuration.

\input{figure/fig_baseline}

\section{Compared Method Details}
In this section, we provide detailed descriptions of the methods and baselines compared with DEVIAS in the experiments.

\subsection{Single-task supervision.}
\noindent\textbf{Naive baselines.} 
We define the baselines with single supervision, either action or scene, without considering debiasing and disentangling as naive baselines.
We have three naive baselines: i) the naive baseline ViT with action supervision denoted as Naive Action ViT,
ii) the naive baseline ViT with action supervision and some data augmentations denoted as Naive Action ViT w/ Aug.,
iii) the naive baseline ViT with scene supervision denoted as Naive Scene ViT.
%
All three baselines are equipped with ViT-Base as the backbone. 
%
%
We summarize the hyperparameters used in \tabref{action_noaug_hyper}, \tabref{action_aug_hyper} and
 \tabref{scene_vit_hyper}.
In the case of Naive Action ViT w/ Aug., we employ mixup~\cite{zhang2017mixup}, cutmix~\cite{yun2019cutmix}, and random erasing~\cite{zhong2020random} augmentations to add some robustness.
The Naive Scene ViT is exactly the same as the Naive Action ViT except that we use pseudo-scene labels instead of action labels.

\noindent\textbf{Scene debiasing methods.} 
We compare DEVIAS with state-of-the-art self-supervised scene debiasing methods: BE~\cite{BE} and FAME~\cite{fame}.
%
%
We employ a ViT-Base as an encoder and use either BE or FAME as a scene-debiasing data augmentation.
BE randomly selects a frame from the same video and mixes it with the other frames, using a weight drawn from a uniform distribution between 0 and 0.3.
FAME extracts a foreground mask and shuffles the background regions among the videos as denoted in \eqnref{Data_augmentation}, with both $\tau$ and $\rho$ parameters set to 0.5. 
We summarize the hyperparameters used in \tabref{be_hyper} and \tabref{fame_hyper}.
%

\subsection{Multi-task supervision.}
We provide a detailed description and figures for the multi-task baselines.
We show the One-Token baseline in \figref{baselines} (c). We train the one-token model with a single classification token from ViT~\cite{vit}. 
The Two-Token baseline uses two distinct learnable tokens: one is for action and another is for the scene, as depicted in \figref{baselines} (a). 
Both the One-Token and Two-Token baselines have separate heads for action and scene classification.
$L_{action}$ is the cross-entropy loss with the ground-truth action label and $L_{scene}$ is the cross-entropy with the pseudo-scene label.

For a more detailed ablation study, we replace the separate classification heads in \figref{baselines} (a) and (c) with a single unified classification head, as shown in \figref{baselines} (b) and (d). 
In models with a unified classification head, we use the disentangling loss, $L_{DE}$, described in the main paper. 
The hyperparameter setting is identical to the setting for the Naive Action ViT.
We summarize the hyperparameters used in \tabref{action_noaug_hyper}.
Additionally, we add either the BE or FAME as a data augmentation on top of the multi-task baseline with separate classification heads (\figref{baselines} (c)). 
We summarize the hyperparameters used for these settings in \tabref{be_hyper} and \tabref{fame_hyper}.
%
%
%


\section{Dataset Details}
In this section, we provide a detailed description of the datasets. 

\subsection{Training dataset}
We train DEVIAS and baselines on the train set of the UCF-101~\cite{soomro2012ucf101}, and Kinetics-400~\cite{kay2017kinetics} datasets. The UCF-101~\cite{soomro2012ucf101} dataset consists of 9,537 training videos and 3,783 test videos from 101 classes.
The Kinetics-400~\cite{kay2017kinetics} dataset comprises $\sim$ 240K training videos and $\sim$20K validation videos from 400 classes. 

%
To evaluate the proposed method using more realistic data, we also train the models on the HVU \cite{diba2020hvu} dataset.
The HVU~\cite{diba2020hvu} provides multiple task annotations including action and scene annotations per video. 
In the original HVU train split, there are 480k videos with 739 action categories and 248 scene categories.
To evaluate the action and scene recognition performance in both the seen and unseen combinations scenarios, we sample 27,532 videos out of the 480k videos from the original train split. 
Each sampled video has a single action label and a single scene label.
In the resulting train split, we have 641 action and 184 scene categories.

\subsection{Test dataset}
For testing, we use the test or validation set of diverse datasets for thorough evaluation: UCF-101, Kinetics-400, HAT~\cite{chung2022hatdataset}, SCUBA~\cite{li2023stillmix}, and HVU~\cite{diba2020hvu}.
Since the other datasets except the HVU dataset do not have ground-truth scene labels, we generate pseudo-scene labels using the scene model described in \secref{devias_detail}.
We categorize the datasets into seen combination datasets, unseen combination datasets, and realistic dataset. 



\noindent\textbf{Seen combination datasets.} 
The \emph{seen combination} dataset consists of action-scene combinations used in the training time: \eg playing basketball on a basketball court.
In action recognition, datasets such as UCF-101~\cite{soomro2012ucf101}, and Kinetics-400~\cite{kay2017kinetics} are considered as \emph{seen combination} datasets, exhibiting a high correlation between action and scene~\cite{li2018resound,whycantchoi}.
%
We evaluate our model on the test set of the UCF-101 dataset (3,783 videos) and the validation set of the Kinetics-400 dataset ($\sim$ 20K videos) in this paper. 

%
%
%
%

\noindent\textbf{Unseen combination datasets.} 
The \emph{unseen combination} dataset consists of action-scene combinations not used during the training time: \eg dancing in the mall~\cite{whycantchoi}.
SCUBA~\cite{li2023stillmix} and HAT~\cite{chung2022hatdataset} synthesize \emph{unseen combination} videos by applying segmentation models to \emph{seen combination} videos such as UCF-101 and Kinetics-400. 
Please note that both the SCUBA and HAT datasets do not provide a training set.

SCUBA~\cite{li2023stillmix} comprises videos created by superimposing action regions extracted from a video onto various scenes. There are three different scene sources: i) the validation set of Places365~\cite{places365}, ii) 2000 scene images generated by VQGAN-CLIP~\cite{crowson2022vqgan} with the template of a random category of Places365 and a random artistic style, and iii) random images with S-shaped stripe patterns generated by sinusoidal functions.
The UCF-101-SCUBA consists of combinations of 910 videos in the first split of the UCF-101 validation set and five randomly selected scene images from each source, resulting in a total of 4,550 videos for each source. 
Each set of the Kinetics-400-SCUBA, \ie Kinetics-400-SCUBA-Places365, Kinetics-400-SCUBA-VQGAN-CLIP, and Kinetics-400-SCUBA-Sin usoid, contains 10,190 videos for each background source, generated by pairing 10,190 videos from the Kinetics-400 validation set with a single background image. 
There exists a version assessed by Amazon Mechanical Turk (AMT) human workers to determine whether the video clearly holds the action. 
%
%
Following the previous work~\cite{li2023stillmix}, we use the version without AMT human assessment for a fair comparison.

HAT~\cite{chung2022hatdataset} is another synthesized dataset for evaluating the effect of scene (background) bias in action recognition models. 
In this dataset, the `Scene-Only' set consists of videos with the background only. The set is generated by removing all human regions and inpainting the human regions. 
The `Scene-Only' set has 19,877 videos in the HAT-Kinetics-400 dataset and 3,783 videos in the HAT-UCF-101 dataset.

The `Action-Swap' set comprises synthesized videos by combining the segmented human regions from a video with a background video generated by human-region inpainting. 
There are four different versions of the `Action-Swap' set: `Random', `Close', `Far', and `Same'.
The `Random' indicates that the background video belongs to a random action class. 
The `Close' and `Far' videos are the result of precisely manipulating the selection of the backgrounds to be either similar to or significantly different from the original background. 
To assess the similarity of the backgrounds between a pair of videos, the authors~\cite{chung2022hatdataset} employ a Places365~\cite{places365} trained scene classification model. 
The `Same' consists of videos where the background comes from a video with the same action class as the original video. 
We exclude the `Same' in our experiments as we use the original UCF-101 and Kinetics datasets as seen combination datasets.
The `Action-Swap' set has 5,631 videos in the HAT-Kinetics-400 dataset. 

Since there are no publicly available annotations of `Random', `Close', `Far', and `Same' sets in the HAT-UCF-101 dataset, we create and employ our own. 
Following the previous work~\cite{chung2022hatdataset}, we use 1,572 videos only where human masks cover between 5\% to 50\% of the total pixel count.
In the HAT-UCF-101 dataset, we use the closest 5 action classes for `Close' and the farthest 30 action classes for `Far'. 
Each version of the `Action-Swap' set has three splits with different combinations of actions and backgrounds. 

\noindent\textbf{Realistic dataset.} 
To evaluate performance using more realistic data, we rearrange the holistic video understanding (HVU) dataset~\cite{diba2020hvu} which provides both action and scene labels.
Since the HVU \cite{diba2020hvu} dataset contains both action and scene labels, we utilize the samples from the validation set to construct seen and unseen combination splits. 
We observe that the HVU is highly imbalanced in terms of both action and scene labels. Therefore, we select the top 400 most frequent action labels and the top 20 most frequent scene labels in the training set, resulting in a total of 1365 videos.
Within this set, the unseen combination split consists of combinations of action labels and scene labels that appear five times or less in the training set, totaling approximately 200 videos. And, the seen combination split is composed of the top 200 videos based on the frequency of occurrence of combinations of action labels and scene labels in the training set.


\section{Evaluation Metric}
To gauge a model’s balanced performance in both seen and unseen combinations scenarios, we report the harmonic mean (H.M.) across four performance metrics as our main metric: i) seen, ii) unseen combinations action, iii) seen, and iv) unseen combinations scene.
The harmonic mean (H.M.) is a widely accepted metric for assessing the \emph{balanced performance} across multiple measures. A well-known example is the F1 score: a harmonic mean of precision and recall. Many works on open-set recognition/adaptation, \eg ANNA~\cite{li2023adjustment}, UADAL~\cite{jang2022unknown}, OVANet~\cite{saito2021ovanet}, and ROS~\cite{bucci2020effectiveness}
also use the H.M. to measure the balanced performance of recognizing seen and unseen classes. Therefore, we use the H.M. to measure the balanced seen/unseen combinations of action/scene performance.

\input{table/supple/ucf101_full}
\input{table/supple/k400_full}
\section{Additional Results}
\label{sec:results}

\subsection{Comprehensive experimental results}
\label{sec:full-results}

To provide comprehensive experimental results, we augment the tables in the main paper by incorporating the \emph{unseen combination} action recognition performances on the HAT~\cite{chung2022hatdataset} dataset. 
We append the results on the HAT `Random', `Close', and `Far' for action recognition and the results on the `Close' for the scene recognition.
We show the arithmetic mean of unseen combination action recognition performance of the six performances on the SCUBA~\cite{li2023stillmix} and HAT~\cite{chung2022hatdataset} datasets: `SCUBA-Places365', `SCUBA-VQGAN-CLIP', `SCUBA-Sinusoid', `HAT-Random', `HAT-Close', and `HAT-Far'.
For the unseen combination scene recognition, we show the arithmetic mean of the four performances of the HAT dataset: `Scene-Only', `Random', `Close', and `Far'. 
To assess the \emph{balanced performance} of i) action \& scene recognition, and ii) seen combination \& unseen combination recognition, we report the harmonic mean (H.M.) of the four performances: seen combination action recognition, unseen combination action recognition (arithmetic mean), seen combination scene recognition, and unseen combination scene recognition (arithmetic mean).
As shown in \tabref{sup_ucf} and \tabref{sup_k400}, DEVIAS achieves a significant improvement of 9.6 points and 1.5 points over the second-best method in terms of the harmonic mean on UCF-101 and Kinetics-400, respectively.

\subsection{Linear probe experimental results}
%
In \tabref{sup_ucf_linear} and \tabref{sup_k400_linear}, we use the linear probe evaluation protocol to measure the overall performance for the single-task baselines for which direct supervision is not provided. 
For instance, after training a model with action supervision only, we evaluate the scene performance using the linear probe protocol. 
Similarly, when we train a model with scene supervision, we apply the linear probe protocol to evaluate the action performance. 
For the scene and action linear probe evaluation, we use the same hyperparameters.
We summarize the hyperparameters used for each dataset in \tabref{lp_hyper}. 
As shown in \tabref{sup_ucf_linear} and \tabref{sup_k400_linear}, DEVIAS shows significant improvements in harmonic mean over the single-task baselines.
%
\input{table/supple/ucf_linear}
\input{table/supple/k400_linear}

\subsection{Comprehensive downstream task results}
\input{table/supple/downstream}

We provide detailed descriptions of the methods and baselines compared with DEVIAS in the downstream experiments.
We employ the Kinetics-400 as our source dataset, and the Diving48~\cite{li2018resound}, Something-Something V2~\cite{goyal2017something}, UCF-101~\cite{soomro2012ucf101}, and ActivityNet~\cite{caba2015activitynet} as our target datasets. 
We choose multiple datasets with distinct characteristics, including the temporal-biased
Diving48 and Something-Something-V2, scene-biased UCF-101 and ActivityNet.
For a quick experiment, we use the first 88 classes (0 to 87) out of the 174 classes of the Something-Something-V2 dataset. 
We compare DEVIAS with the single-task baselines (Naive Action ViT, Naive Scene ViT, BE~\cite{BE}, FAME~\cite{fame}), and the multi-task baseline (Two-Token, Two-Token w/ FAME) using the full fine-tuning protocol.
For the multi-task baseline and DEVIAS, we train the downstream tasks using two approaches. First, we aggregate the features passed through the final block of the backbone encoder by average pooling across the temporal and spatial axes to yield a single feature vector. 
We feed the feature vector into the classification head, which we denote as GAP.
Secondly, we project the action and scene tokens to half dimension and concatenate both tokens. And we feed them into the classification head, which is denoted as concat..

We show the results in \tabref{downstream} and summarize the hyperparameters used in \tabref{downstream_hyper}.
Compared to the baselines, both DEVIAS (GAP) and (concat.) shows favorable performance on the downstream tasks across the temporal-biased and the scene-biased tasks.
%
We expect enhanced performances on the downstream tasks with more advanced fusion methods.
We leave the investigation on more advanced fusion methods as our future work.
%
%
\input{table/supple/scene_model}

\subsection{Experiments using different scene models.}
\label{sec:scene_model_exp}
In \tabref{scene_model}, we study the effect of using different pre-trained scene models to generate pseudo scene labels. 
In addition to the Places365 \cite{places365} pre-trained model used in the main paper, we also conduct experiments with the SUN397 \cite{sun} pre-trained model and CLIP \cite{radford2021clip}. To obtain pseudo scene labels using CLIP, we provide CLIP with the center frame as input and perform zero-shot classification using the scene labels from Places365.
For evaluating action performance, we report top-1 accuracy on the validation split of UCF-101 (seen), and SCUBA-VQGAN-CLIP~\cite{li2023stillmix} (unseen). For scene performance, we report top-5 accuracy on the validation split of UCF-101 (seen) and UCF-101-Scene-only~\cite{chung2022hatdataset} (unseen).
And we obtain pseudo scene labels for testing using the same scene model used in the training phase.
As we observe in \tabref{scene_model}, DEVIAS outperforms other baselines in multi-task learning using scene models pre-trained on various datasets. This indicates that DEVIAS is a robust approach across different types of scene models.
%
\input{table/supple/unified}

\subsection{Effect of unified classification head}
In \tabref{unified}, we study the effect of the unified classification head. 
We replace the separate classification heads in One-Token and Two-Token baselines with a single unified classification head denoted as One-Token (unified head) and Two-Token (unified head).
%
%
Using the unified classification head on top of the Two-Token baseline results in a performance improvement compared to using the separate classification heads.
Therefore, we employ the unified classification head in our DEVIAS throughout the paper.

\input{table/supple/ablation_full}
\subsection{Comprehensive ablation results}
\noindent\textbf{Effect of the cosine similarity loss.}
We investigate the effect of the cosine similarity loss in the \tabref{abl} (a). Applying the cosine similarity loss, results in a gain of 0.5 points in the harmonic mean.

\noindent\textbf{Effect of the action mask decoder.}
In \tabref{abl} (b), we analyze the effect of the action mask decoder (AMD). When applying a mask predictor loss, we observe an increase of 1.6 points in the harmonic mean performance. 
When we use attention guidance, it further enhances the performance by 5.5 points in the harmonic mean. 
Using both methods together results in a performance improvement of 5.9 points in the harmonic mean. 
Notably, when we apply attention guidance and AMD without the disentangling encoder, performance decreases. 
This result demonstrates the crucial importance of using both the disentangling encoder and the action mask decoder for effective disentangle representation learning.


\noindent\textbf{Ablation study on backbone architecture.}
To validate the effectiveness of DEVIAS when equipped with a CNN backbone architecture, we employ I3D~\cite{carreira2018i3d} as a backbone encoder. 
As shown in \tabref{abl} (c), DEVIAS outperforms baselines with significant margins, demonstrating its compatibility with various backbone architectures.

\noindent\textbf{Ablation study on pre-trained weights.}
To study the effect of using different pre-training strategies, we show the results of using Omnivore~\cite{girdhar2022omnivore} pre-trained weights, in \tabref{abl} (d). 
DEVIAS outperforms baselines with significant margins, showing that the performance superiority of our method does not depend on the pre-trained weights.




\subsection{Slot assignment}
DEVIAS utilizes the Hungarian algorithm~\cite{kuhn1955hungarian} as a matching function during the training.
During the inference, DEVIAS uses the softmax probability-based slot assignment.
%
To validate whether the learned action and scene slots take the same role during the inference or not, we measure the slot assignment frequency of slot 1 and slot 2, \ie the frequency of the slot assignment as either the action slot or the scene slot.
We measure the slot assignment frequency on the original UCF-101 (seen combination), UCF-101-SCUBA-VQGAN-CLIP, UCF-101-Scene-Only, and UCF-101-HAT-Far (unseen combination) datasets. 
Since all results are exactly the same as \figref{slot_assign}, we only show the results on the UCF-101-SCUBA-VQGAN-CLIP dataset. 
As we observe in \figref{slot_assign}, the softmax probability-based assignment assigns slot 1 as the action slot with $100\%$ frequency and slot 2 as the scene slot with $100\%$ frequency.
%

\input{figure/supple/fig_slot_assignment}

\input{figure/fig_umap}

\input{figure/supple/rebuttal/fig_attnmap}


\subsection{Qualitative results}
For a more comprehensive understanding, we conduct three qualitative analyses. we first show the slot attention map in \figref{full_slot_viz}.
We select samples from the validation set of the two datasets: UCF-101 for seen combination recognition and `HAT-Far' for unseen combination senarios. 
In both cases, each slot clearly focuses on its designated region.
We also show UMAP~\cite{mcinnes2018umap} visualization of the feature vectors of Two-Token, Two-Token w/ BE, and Two-Token w/ FAME, and DEVIAS as shown in \figref{umap}. Compared to the baselines, DEVIAS clearly demonstrates a distinct separation between action and scene feature vectors.
To demonstrate the efficacy of the action mask decoder (AMD), we visualize the attention map of the action slot in \figref{attnmap}. 
When using AMD, the action slot accurately focuses on the action regions, indicating the effective learning of disentangled action and scene representations.

\input{figure/supple/fig_full_slot_viz}
\input{table/supple/scene_hyper}
\input{table/supple/action_noaug_hyper}

\input{table/supple/action_aug_hyper}

\input{table/supple/scene_vit_hyper}

\input{table/supple/be}

\input{table/supple/fame}

\input{table/supple/linearprobe_hyper}

\input{table/supple/hyperpramas}

\input{table/supple/downstream_hyper}

%% file: figure/fig_baseline.tex
\begin{figure}[t]
    \centering
    \includegraphics[width=1\linewidth]{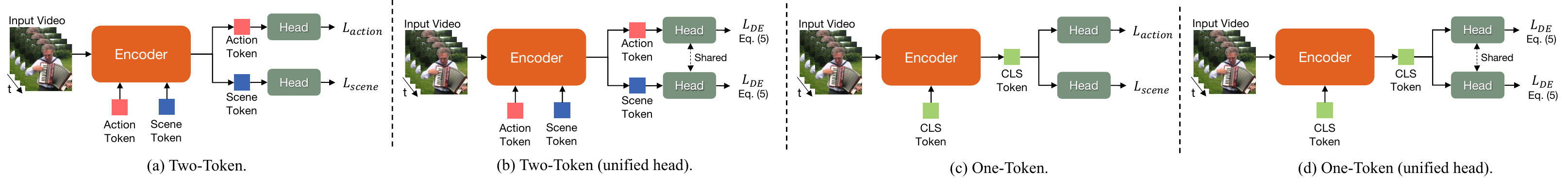} 
    
    \figcapmargin
    \caption{\textbf{Architecture of baselines.} 
    All baselines use the same encoder, ViT~\cite{vit}. (a) and (b) use separate tokens for action and scene, and (c) and (d) use a single token. (a) and (c) use separate classification heads for action and scene, and (b) and (d) use a unified classification head.
    }
    \figcapmargin    
    \label{fig:baselines}
\end{figure}

%% file: table/supple/ucf101_full.tex
\begin{table*}[t]
\centering
\caption{\textbf{Action and scene recognition performance on the UCF-101 dataset.} We report the Top-1 action recognition accuracy and the Top-5 scene recognition accuracy. We evaluate both seen and unseen recognition performances. We also report the harmonic mean (H.M.) of the action recognition and scene recognition. V.C./Sin. denotes the SCUBA~\cite{li2023stillmix} VQGAN-CLIP/Sinusoidal; S.O./Rand. denotes the HAT~\cite{chung2022hatdataset} Scene-Only/Random. The \best{best} and the  \second{second-best} H.M. numbers are highlighted.}

\def\arraystretch{1.2}

\resizebox{1.0\linewidth}{!}{
\begin{tabular}{ lccl c ccc c ccc   c   c  cccc ccc}
\toprule

\multirowcell{3}[-1.5ex]{\shortstack{Training \\ Strategy}} & \multicolumn{2}{c}{\multirowcell{3}[-0.7ex]{Supervision}} &  & \multicolumn{9}{c}{Action($\uparrow$) } & \multicolumn{7}{c}{Scene($\uparrow$)} & \\

\cline{5-13}
\cline{15-20}
 && &   & \multirow{3}{*}{Seen} & \multicolumn{8}{c}{Unseen} && \multirow{3}{*}{Seen} & \multicolumn{4}{c}{Unseen}  && \\
\cline{6-13} \cline{16-20}
&&&\multicolumn{1}{c}{Method}&& \multicolumn{3}{c}{\multirowcell{1}[-1.0ex]{SCUBA~\cite{li2023stillmix}}} && \multicolumn{3}{c}{\multirowcell{1}[-1.0ex]{HAT~\cite{chung2022hatdataset}}} & \multirow{2}{*}{Mean} &&& \multicolumn{4}{c}{\multirowcell{1}[-1.0ex]{HAT~\cite{chung2022hatdataset}}} & \multirow{2}{*}{Mean} & \\
\cmidrule{2-3} \cline{6-8} \cline{10-12} \cline{16-19}
&Action&Scene & & & Places365 & V.C.  & Sin. && Rand. & Close & \multicolumn{1}{c}{Far} & &&& S.O.& Rand. & Close & Far & & \multirow{-4}{*}{H.M.} \\

\midrule
      
\multirow{5}{*}{Single-Task} &\checkmark&$\times$&  Naive Action ViT   & \cellcolor{gray!30}92.9 & 15.0 & 12.4 & 21.0 && 23.5 & 27.7 & 14.1 &\cellcolor{gray!30}19.0 &&\cellcolor{gray!30}-&-&-&-&-&\cellcolor{gray!30}-&\cellcolor{gray!30}-\\
&\checkmark&$\times$ &\shortstack{Action ViT w/ Aug. } & \cellcolor{gray!30}90.0 & 19.1 & 18.8 & 19.4 && 21.4 & 27.3 & 13.0 &\cellcolor{gray!30}19.8 &&\cellcolor{gray!30}-&-&-&-&-&\cellcolor{gray!30}-&\cellcolor{gray!30}-\\
&\checkmark&$\times$ & BE \cite{BE}  & \cellcolor{gray!30}92.3 & 16.1 & 12.1 & 38.7 && 29.6 & 34.5 & 19.2 &\cellcolor{gray!30}25.0& &\cellcolor{gray!30}-&-&-&-&-&\cellcolor{gray!30}-&\cellcolor{gray!30}-\\
 &\checkmark&$\times$ & FAME \cite{fame}  & \cellcolor{gray!30}91.6 & 22.0 & 24.8 & 15.6 && 36.7 & 41.7 & 30.6 &\cellcolor{gray!30}28.6 &&\cellcolor{gray!30}-&-&-&-&-&\cellcolor{gray!30}-&\cellcolor{gray!30}-\\

&$\times$&\checkmark & Naive Scene ViT  &\cellcolor{gray!30}-&-&-&-&-&-&- &-&\cellcolor{gray!30}-&& \cellcolor{gray!30}72.0 & 61.7 & 62.8 & 60.8 & 69.6 & \cellcolor{gray!30}63.7& \cellcolor{gray!30}- \\

\midrule

\multirow{4}{*}{Multi-Task} &\checkmark&\checkmark &  One-Token & \cellcolor{gray!30}91.9 & 10.5 & 5.0 & 21.8 && 19.8 & 27.9 & 8.8 & \cellcolor{gray!30}15.6 & & \cellcolor{gray!30}74.0 & 60.5 & 58.0 & 57.6 & 66.5 & \cellcolor{gray!30}60.7 & \cellcolor{gray!30}38.1 \\
 &\checkmark&\checkmark  &  Two-Token & \cellcolor{gray!30}86.0 & 11.9 & 11.1 & 19.9 && 17.3 & 23.9 & 9.1 &\cellcolor{gray!30}15.5 && \cellcolor{gray!30}72.3 & 59.6 & 59.2 & 57.8 & 67.1 &\cellcolor{gray!30}60.9 &  \cellcolor{gray!30}37.6\\
  &\checkmark&\checkmark & Two-Token w/ BE \cite{BE} & \cellcolor{gray!30}89.9 & 15.0 & 13.0 & 20.5 && 22.4 & 29.0 & 12.2 & \cellcolor{gray!30}18.7 & & \cellcolor{gray!30}74.2 & 62.3 & 59.3 & 58.4 & 69.5 &\cellcolor{gray!30}62.4 &\cellcolor{gray!30}42.5 \\
&\checkmark&\checkmark &  Two-Token w/ FAME \cite{fame} & \cellcolor{gray!30}89.5 & 21.7 & 25.3 & 15.3 && 32.9 & 38.2 & 26.4 & \cellcolor{gray!30}26.6 && \cellcolor{gray!30}73.2 & 61.4 & 62.8 & 61.3 & 70.3 & \cellcolor{gray!30}64.0 & \cellcolor{gray!30}\second{51.2}\\
\midrule
 Disentangle& \checkmark&\checkmark& DEVIAS & \cellcolor{gray!30}90.1 & 41.1 & 40.1 & 38.6 && 39.4 & 41.3 & 35.2 &\cellcolor{gray!30}39.3&& \cellcolor{gray!30}74.0 & 61.0 & 62.4 & 59.8 & 70.2 &\cellcolor{gray!30}63.4& \cellcolor{gray!30}\best{60.8}\\

\bottomrule
\end{tabular}

}
\label{tab:sup_ucf}
\end{table*}

%% file: table/supple/k400_full.tex
\begin{table*}[t]
\centering
\caption{\textbf{Action and scene recognition performance on the Kinetics-400 dataset.} We report the Top-1 action recognition accuracy and the Top-5 scene recognition accuracy. We evaluate both seen and unseen recognition performances. We also report the harmonic mean (H.M.) of the action recognition and scene recognition. V.C./Sin. denotes the SCUBA~\cite{li2023stillmix} VQGAN-CLIP/Sinusoidal; S.O./Rand. denotes the HAT~\cite{chung2022hatdataset} Scene-Only/Random. The \best{best} and the \second{second-best} H.M. numbers are highlighted.
}

\def\arraystretch{1.2}

\resizebox{1.0\linewidth}{!}{
\begin{tabular}{ lccl c ccc c ccc   c   c  cccc ccc}
\toprule

\multirowcell{3}[-1.5ex]{\shortstack{Training \\ Strategy}} & \multicolumn{2}{c}{\multirowcell{3}[-0.7ex]{Supervision}} &  & \multicolumn{9}{c}{Action($\uparrow$) } & \multicolumn{7}{c}{Scene($\uparrow$)} & \\

\cline{5-13}
\cline{15-20}
 && &   & \multirow{3}{*}{Seen} & \multicolumn{8}{c}{Unseen} && \multirow{3}{*}{Seen} & \multicolumn{4}{c}{Unseen}  && \\
\cline{6-13} \cline{16-20}
&&&\multicolumn{1}{c}{Method}&& \multicolumn{3}{c}{\multirowcell{1}[-1.0ex]{SCUBA~\cite{li2023stillmix}}} && \multicolumn{3}{c}{\multirowcell{1}[-1.0ex]{HAT~\cite{chung2022hatdataset}}} & \multirow{2}{*}{Mean} &&& \multicolumn{4}{c}{\multirowcell{1}[-1.0ex]{HAT~\cite{chung2022hatdataset}}} & \multirow{2}{*}{Mean} & \\
\cmidrule{2-3} \cline{6-8} \cline{10-12} \cline{16-19}
&Action&Scene & & & Places365 & V.C.  & Sin. && Rand. & Close & \multicolumn{1}{c}{Far} & &&& S.O.& Rand. & Close & Far & & \multirow{-4}{*}{H.M.} \\

\midrule
      
\multirow{5}{*}{Single-Task} & \checkmark\ &$\times$& Naive Action ViT    & \cellcolor{gray!30}76.8 & 41.3 &  41.6 & 49.6 && 14.4 & 23.2 & 11.4 &\cellcolor{gray!30}30.3 &&\cellcolor{gray!30}-&-&-&-&-&\cellcolor{gray!30}-&\cellcolor{gray!30}-\\
&\checkmark\ &$\times$&  \shortstack{Naive Action ViT w/ Aug.}   & \cellcolor{gray!30}77.6 & 50.7 &  49.4 &\multicolumn{1}{c}{57.3} && 15.1 & 24.7 & 11.9  &\cellcolor{gray!30}34.9 &&\cellcolor{gray!30}-&-&-&-&-&\cellcolor{gray!30}-&\cellcolor{gray!30}-\\
 &\checkmark\ &$\times$& BE \cite{BE}   & \cellcolor{gray!30}77.6 & 43.1 & 43.2 & \multicolumn{1}{c}{52.2} && 15.1 & 24.1 & 11.7 & \cellcolor{gray!30}31.6 & &\cellcolor{gray!30}-&-&-&-&-&\cellcolor{gray!30}-&\cellcolor{gray!30}- \\
 &\checkmark\ &$\times$&   FAME \cite{fame}   & \cellcolor{gray!30}77.8 & 49.3  & 49.7  & \multicolumn{1}{c}{56.8} && 22.9 & 29.6 & 19.9 &\cellcolor{gray!30}38.0 &&\cellcolor{gray!30}-&-&-&-&-&\cellcolor{gray!30}-&\cellcolor{gray!30}-\\


&$\times$&\checkmark\   & Naive Scene ViT   &\cellcolor{gray!30}-&-&-&-&-&-&-&-&\cellcolor{gray!30}- && \cellcolor{gray!30}86.5 & 82.6 & 79.9 & 81.0 & 81.2 & \cellcolor{gray!30}81.2 & \cellcolor{gray!30}- \\

\midrule

\multirow{4}{*}{Multi-Task} &\checkmark&\checkmark& One-Token & \cellcolor{gray!30}74.2 & 34.5 & 35.2 & \multicolumn{1}{c}{45.6} && 11.9 & 21.3 & 8.7 & \cellcolor{gray!30}26.2& & \cellcolor{gray!30}87.9 & \multicolumn{1}{c}{83.8} & 80.8 & 82.7 & 81.5 & \cellcolor{gray!30}82.2 & \cellcolor{gray!30}53.2 \\
&\checkmark&\checkmark&  Two-Token & \cellcolor{gray!30}75.1 & 35.3 & 34.9 & \multicolumn{1}{c}{46.6} && 12.5 & 21.5 & 9.3 &\cellcolor{gray!30}26.7 && \cellcolor{gray!30}86.4 & 75.8 & 78.3 & 79.4 & 80.3 &\cellcolor{gray!30}78.5&  \cellcolor{gray!30}53.3\\
  &\checkmark&\checkmark& Two-Token w/ BE \cite{BE} & \cellcolor{gray!30}75.1 & 35.7 & 35.8 & 48.0 && 13.0 & 21.9 & 9.7 & \cellcolor{gray!30}27.4& & \cellcolor{gray!30}87.7 & 82.9 & 80.0 & 81.4 & 81.5 &\cellcolor{gray!30}81.5 &\cellcolor{gray!30}54.4 \\
&\checkmark&\checkmark& Two-Token w/ FAME \cite{fame} & \cellcolor{gray!30}75.0 & 43.0 & 45.5 &\multicolumn{1}{c}{51.5} && 20.7 & 28.2 & 17.9 & \cellcolor{gray!30}34.5 && \cellcolor{gray!30}87.3 & 77.4 & 81.1 & 81.7 & 82.6 & \cellcolor{gray!30}80.7 & \cellcolor{gray!30}\second{60.5} \\
\midrule
 Disentangle&\checkmark&\checkmark& DEVIAS & \cellcolor{gray!30}77.3 & 48.9 & 50.3 & \multicolumn{1}{c}{58.8} && 21.8 & 30.1 & 18.9 &\cellcolor{gray!30}38.1 && \cellcolor{gray!30}82.0 & 76.5 & 75.7 & 76.0 & 77.1 &\cellcolor{gray!30}76.3 & \cellcolor{gray!30}\best{62.0}\\

\bottomrule
\end{tabular}

}
\label{tab:sup_k400}
\end{table*}

%% file: table/supple/ucf_linear.tex
\begin{table*}[t]
\centering
\caption{\textbf{Action and scene recognition performance on the UCF-101 dataset.} We report the Top-1 action recognition accuracy and the Top-5 scene recognition accuracy. We evaluate both seen and unseen recognition performances. We also report the harmonic mean (H.M.) of the action recognition and scene recognition. V.C./Sin. denotes the SCUBA~\cite{li2023stillmix} VQGAN-CLIP/Sinusoidal; S.O./Rand. denotes the HAT~\cite{chung2022hatdataset} Scene-Only/Random. $\dagger$ indicates that we use a linear probe evaluation for the task for which direct supervision is not provided. The \best{best} and the  \second{second-best} H.M. numbers are highlighted.}

\def\arraystretch{1.2}

\resizebox{1.0\linewidth}{!}{
\begin{tabular}{ lccl c ccc c ccc   c   c  cccc ccc}
\toprule

\multirowcell{3}[-1.5ex]{\shortstack{Training \\ Strategy}} & \multicolumn{2}{c}{\multirowcell{3}[-0.7ex]{Supervision}} &  & \multicolumn{9}{c}{Action($\uparrow$) } & \multicolumn{7}{c}{Scene($\uparrow$)} & \\

\cline{5-13}
\cline{15-20}
 && &   & \multirow{3}{*}{Seen} & \multicolumn{8}{c}{Unseen} && \multirow{3}{*}{Seen} & \multicolumn{4}{c}{Unseen}  && \\
\cline{6-13} \cline{16-20}
&&&\multicolumn{1}{c}{Method}&& \multicolumn{3}{c}{\multirowcell{1}[-1.0ex]{SCUBA~\cite{li2023stillmix}}} && \multicolumn{3}{c}{\multirowcell{1}[-1.0ex]{HAT~\cite{chung2022hatdataset}}} & \multirow{2}{*}{Mean} &&& \multicolumn{4}{c}{\multirowcell{1}[-1.0ex]{HAT~\cite{chung2022hatdataset}}} & \multirow{2}{*}{Mean} & \\
\cmidrule{2-3} \cline{6-8} \cline{10-12} \cline{16-19}
&Action&Scene & & & Places365 & V.C.  & Sin. && Rand. & Close & \multicolumn{1}{c}{Far} & &&& S.O.& Rand. & Close & Far & & \multirow{-4}{*}{H.M.} \\

\midrule
      
\multirow{5}{*}{Single-Task$^\dagger$ } &\checkmark&$\times$&  Naive Action ViT   & \cellcolor{gray!30}92.9 & 15.0 & 12.4 & 21.0 && 23.5 & 27.7 & 14.1 &\cellcolor{gray!30}19.0 &&\cellcolor{gray!30}62.9&   50.2 &  49.9 & 50.8 & 56.6  & \cellcolor{gray!30}51.9& \cellcolor{gray!30}40.6\\
&\checkmark&$\times$ &\shortstack{Action ViT w/ Aug.} & \cellcolor{gray!30}90.0 & 19.1 & 18.8 & 19.4 && 21.4 & 27.3 & 13.0 &\cellcolor{gray!30}19.8 && \cellcolor{gray!30}64.0 & 52.0 & 54.0 & 51.9 & 63.3 &  \cellcolor{gray!30}55.3& \cellcolor{gray!30}42.0\\
&\checkmark&$\times$ & BE \cite{BE} & \cellcolor{gray!30}92.3 & 16.1 & 12.1 & 38.7 && 29.6 & 34.5 & 19.2 &\cellcolor{gray!30}25.0& &\cellcolor{gray!30}63.6 & 51.7 & 52.6 & 52.1 & 59.4 &\cellcolor{gray!30}54.0 & \cellcolor{gray!30}47.0\\
 &\checkmark&$\times$ & FAME \cite{fame} & \cellcolor{gray!30}91.6 & 22.0 & 24.8 & 15.6 && 36.7 & 41.7 & 30.6 &\cellcolor{gray!30}28.6 && \cellcolor{gray!30}62.7 & 51.2 & 48.7 & 50.8 & 54.5 &\cellcolor{gray!30}51.3 & \cellcolor{gray!30}\second{49.2}\\

&$\times$&\checkmark & Naive Scene ViT  & \cellcolor{gray!30}69.2 & 2.2 & 0.9 & 7.1 && 8.7 & 14.3 & 3.2 & \cellcolor{gray!30}6.1 && \cellcolor{gray!30}72.0 & 61.7 & 62.8 & 60.8 & 69.6 & \cellcolor{gray!30}63.7& \cellcolor{gray!30}19.2 \\

\midrule

 Disentangle& \checkmark&\checkmark& DEVIAS & \cellcolor{gray!30}90.1 & 41.1 & 40.1 & 38.6 && 39.4 & 41.3 & 35.2 &\cellcolor{gray!30}39.3&& \cellcolor{gray!30}74.0 & 61.0 & 62.4 & 59.8 & 70.2 &\cellcolor{gray!30}63.4& \cellcolor{gray!30}\best{60.8}\\

\bottomrule
\end{tabular}

}
\label{tab:sup_ucf_linear}
\end{table*}

%% file: table/supple/k400_linear.tex
\begin{table*}[t]
\centering
\caption{\textbf{Action and scene recognition performance on the Kinetics-400 dataset.} We report the Top-1 action recognition accuracy and the Top-5 scene recognition accuracy. We evaluate both seen and unseen recognition performances. We also report the harmonic mean (H.M.) of the action recognition and scene recognition. V.C./Sin. denotes the SCUBA~\cite{li2023stillmix} VQGAN-CLIP/Sinusoidal; S.O./Rand. denotes the HAT~\cite{chung2022hatdataset} Scene-Only/Random. $\dagger$ indicates that we use a linear probe evaluation for the task for which direct supervision is not provided. The \best{best} and the  \second{second-best} H.M. numbers are highlighted.
}

\def\arraystretch{1.2}

\resizebox{1.0\linewidth}{!}{
\begin{tabular}{ lccl c ccc c ccc   c   c  cccc ccc}
\toprule

\multirowcell{3}[-1.5ex]{\shortstack{Training \\ Strategy}} & \multicolumn{2}{c}{\multirowcell{3}[-0.7ex]{Supervision}} &  & \multicolumn{9}{c}{Action($\uparrow$) } & \multicolumn{7}{c}{Scene($\uparrow$)} & \\

\cline{5-13}
\cline{15-20}
 && &   & \multirow{3}{*}{Seen} & \multicolumn{8}{c}{Unseen} && \multirow{3}{*}{Seen} & \multicolumn{4}{c}{Unseen}  && \\
\cline{6-13} \cline{16-20}
&&&\multicolumn{1}{c}{Method}&& \multicolumn{3}{c}{\multirowcell{1}[-1.0ex]{SCUBA~\cite{li2023stillmix}}} && \multicolumn{3}{c}{\multirowcell{1}[-1.0ex]{HAT~\cite{chung2022hatdataset}}} & \multirow{2}{*}{Mean} &&& \multicolumn{4}{c}{\multirowcell{1}[-1.0ex]{HAT~\cite{chung2022hatdataset}}} & \multirow{2}{*}{Mean} & \\
\cmidrule{2-3} \cline{6-8} \cline{10-12} \cline{16-19}
&Action&Scene & & & Places365 & V.C.  & Sin. && Rand. & Close & \multicolumn{1}{c}{Far} & &&& S.O.& Rand. & Close & Far & & \multirow{-4}{*}{H.M.} \\

\midrule
      
\multirow{5}{*}{Single-Task$^\dagger$ } & \checkmark\ &$\times$& Naive Action ViT  & \cellcolor{gray!30}76.8 & 41.3 &  41.6 & 49.6 && 14.4 & 23.2 & 11.4 &\cellcolor{gray!30}30.3 && \cellcolor{gray!30}71.2& 65.8& 63.0 & 66.4 & 66.2 & \cellcolor{gray!30}65.4 & \cellcolor{gray!30}53.1 \\
&\checkmark\ &$\times$&  \shortstack{Naive Action ViT w/ Aug.} & \cellcolor{gray!30}77.6 & 50.7 &  49.4 &\multicolumn{1}{c}{57.3} && 15.1 & 24.7 & 11.9  &\cellcolor{gray!30}34.9 && \cellcolor{gray!30}71.6 & 65.7 & 63.7 & 65.5 & 66.3 & \cellcolor{gray!30}65.3 & \cellcolor{gray!30}56.5\\
&\checkmark\ &$\times$& BE \cite{BE} & \cellcolor{gray!30}77.6 & 43.1 & 43.2 & \multicolumn{1}{c}{52.2} && 15.1 & 24.1 & 11.7 & \cellcolor{gray!30}31.6 & & \cellcolor{gray!30}70.7 & 65.4 & 63.0 & 63.7 & 65.0 &\cellcolor{gray!30}64.3 & \cellcolor{gray!30}53.9 \\
 &\checkmark\ &$\times$&   FAME \cite{fame} & \cellcolor{gray!30}77.8 & 49.3  & 49.7  & \multicolumn{1}{c}{56.8} && 22.9 & 29.6 & 19.9 &\cellcolor{gray!30}38.0 && \cellcolor{gray!30}70.3 & 64.9 & 61.0 & 63.3 & 63.4 &\cellcolor{gray!30}63.2& \cellcolor{gray!30}\second{57.8}\\


&$\times$&\checkmark\   & Naive Scene ViT & \cellcolor{gray!30}43.0 & 9.0 & 7.7 & \multicolumn{1}{c}{14.4} && 3.0 & 9.4 & 1.8 & \cellcolor{gray!30}7.6 && \cellcolor{gray!30}86.5 & 82.6 & 79.9 & 81.0 & 81.2 & \cellcolor{gray!30}81.2 & \cellcolor{gray!30}22.4 \\


\midrule
 Disentangle&\checkmark&\checkmark& DEVIAS & \cellcolor{gray!30}77.3 & 48.9 & 50.3 & \multicolumn{1}{c}{58.8} && 21.8 & 30.1 & 18.9 &\cellcolor{gray!30}38.1 && \cellcolor{gray!30}82.0 & 76.5 & 75.7 & 76.0 & 77.1 &\cellcolor{gray!30}76.3 & \cellcolor{gray!30}\best{62.0}\\

\bottomrule
\end{tabular}

}
\label{tab:sup_k400_linear}
\end{table*}

%% file: table/supple/downstream.tex
\begin{table*}[t]
\centering
\caption{\tb{Downstream task performance.} We report Top-1 accuracy (\%). All models are pre-trained on the Kinetics-400 and then fine-tuned on the downstream datasets. SSV2 denotes the Something-Something-V2 dataset. The \best{best} and the \second{second-best} H.M. numbers are highlighted. 
}

\resizebox{0.85\linewidth}{!}{
\begin{tabular}{lll cc c cc c}
    
\toprule
          \multirowcell{2}{\shortstack{Pretraining \\ Strategy}} && \multicolumn{1}{c}{\multirow{2}{*}{Method}}&\multicolumn{2}{c}{Temporal-biased}&&\multicolumn{2}{c}{Scene-biased} &\\
\Xcline{4-5}{0.03em}
\Xcline{7-8}{0.03em}
 &&& Diving48 & SSV2&  & UCF-101 & ActivityNet&\multirow{-2}{*}{H.M.}\\
\midrule
\multirow{4}{*}{Single-Task}&&Naive Action ViT & 81.5 & 74.2&&98.5&84.4&\cellcolor{gray!30}83.8 \\
&&BE~\cite{BE} &81.9 &74.5&& 98.3&84.6&\cellcolor{gray!30}84.0\\
&&FAME~\cite{fame} & 80.6 & 74.2&&98.3&83.8&\cellcolor{gray!30}83.4\\
&&Naive Scene ViT & 73.1 & 71.8&&92.0&73.1&\cellcolor{gray!30}76.7\\
\midrule
\multirow{4}{*}{Multi-Task}&&Two-Token (GAP) & 80.1 & 73.7 && 98.2 & 83.7 & \cellcolor{gray!30}83.0 \\
&&Two-Token (concat.) & 80.1 & 72.4 && 98.3 & 83.2 & \cellcolor{gray!30}82.5 \\
&&Two-Token w/ FAME~\cite{fame} (GAP) & 78.7 & 73.5 && 98.1 & 81.5 & \cellcolor{gray!30}82.0 \\
&&Two-Token w/ FAME~\cite{fame} (concat.) & 78.6 & 71.6 && 98.0 & 82.4 & \cellcolor{gray!30}81.6 \\
\midrule
\multirow{2}{*}{Disentangle}&&DEVIAS (GAP) & 83.1 & 75.3 && 97.8 & 83.1 & \cellcolor{gray!30}\second{84.1} \\
&&DEVIAS (concat.) & 84.4 & 75.2 && 98.4 & 84.5 & \cellcolor{gray!30}\textbf{84.8} \\

\bottomrule
\end{tabular}
}
\label{tab:downstream}
\label{tab:action_noaug_hyper}
\end{table*}

%% file: table/supple/scene_model.tex
\begin{table*}[t]
\centering
\caption{\textbf{Scene model ablation study.}
To evaluate the robustness of using different scene models, we show the results on the UCF-101 dataset. In every experiment, we use a ViT backbone pre-trained on the UCF-101. We report the Top-1 accuracy (\%) for the action and the Top-5 accuracy (\%) for the scene recognition, along with the harmonic mean (H.M.) of the action and scene recognition performances. The \best{best} H.M. numbers are highlighted.
} 

\resizebox{0.8\linewidth}{!}{
    \setlength{\tabcolsep}{3pt}
    \begin{tabular}{ll ccccc}
        \toprule
        \multirow{2}{*}{Pre-train} & \multicolumn{1}{c}{\multirow{2}{*}{Method}}& \multicolumn{2}{c}{Action($\uparrow$)} & \multicolumn{2}{c}{Scene($\uparrow$)} & \multirow{2}{*}{H.M.}\\
        \cmidrule(lr){3-4} \cmidrule(lr){5-6} 
       &&  Seen & Unseen & Seen& Unseen \\
        \midrule
        \multirow{3}{*}{Places365~\cite{places365}} &Two-Token&  86.0 & 11.1 & 72.3 & 59.6 & \cellcolor{gray!30}30.2 \\
         & Two-Token w/ FAME \cite{fame} & 89.5 & 25.3 & 73.2 & 61.4 & \cellcolor{gray!30}49.6\\
         & DEVIAS & 90.1 & 40.1  & 74.0 & 61.0 & \cellcolor{gray!30}\best{60.7}\\
        \midrule
        \multirow{3}{*}{SUN397~\cite{sun}} & Two-Token & 89.2 & 11.5 & 73.3 & 71.0 & \cellcolor{gray!30}31.8\\
        & Two-Token w/ FAME \cite{fame} & 89.4 & 23.5 & 72.0 & 69.1 & \cellcolor{gray!30}48.7\\
         &DEVIAS& 90.0 & 39.9 & 75.0 & 73.3 & \cellcolor{gray!30}\best{63.3}\\
         \midrule
        \multirow{3}{*}{CLIP~\cite{radford2021clip}} & Two-Token & 90.1 & 13.2 & 84.2 & 58.4 & \cellcolor{gray!30}34.5 \\
        & Two-Token w/ FAME \cite{fame} & 89.0 & 24.5 & 84.0 & 57.9 & \cellcolor{gray!30}49.2 \\
         &DEVIAS& 89.3 & 37.4 & 83.9 & 57.5 & \cellcolor{gray!30}\best{59.5}\\
       \bottomrule
    \end{tabular}
}

    
\label{tab:scene_model}
\end{table*}

%% file: table/supple/unified.tex
\begin{table*}[t]
\centering
\caption{\textbf{Effect of unified classification head on the UCF-101 dataset.} We report the Top-1 action recognition accuracy (\%) and the Top-5 scene recognition accuracy (\%). We evaluate both seen and unseen recognition performances. We also report the harmonic mean (H.M.) of the action recognition and scene recognition performances. V.C./Sin. denotes the SCUBA~\cite{li2023stillmix} VQGAN-CLIP/Sinusoidal; S.O./Rand. denotes the HAT~\cite{chung2022hatdataset} Scene-Only/Random.}

\def\arraystretch{1.2}

\resizebox{1.0\linewidth}{!}{
\begin{tabular}{ lccl c ccc c ccc   c   c  cccc ccc}
\toprule

\multirowcell{3}[-1.5ex]{\shortstack{Training \\ Strategy}} & \multicolumn{2}{c}{\multirowcell{3}[-0.7ex]{Supervision}} &  & \multicolumn{9}{c}{Action($\uparrow$) } & \multicolumn{7}{c}{Scene($\uparrow$)} & \\

\cline{5-13}
\cline{15-20}
 && &   & \multirow{3}{*}{Seen} & \multicolumn{8}{c}{Unseen} && \multirow{3}{*}{Seen} & \multicolumn{4}{c}{Unseen}  && \\
\cline{6-13} \cline{16-20}
&&&\multicolumn{1}{c}{Method}&& \multicolumn{3}{c}{\multirowcell{1}[-1.0ex]{SCUBA~\cite{li2023stillmix}}} && \multicolumn{3}{c}{\multirowcell{1}[-1.0ex]{HAT~\cite{chung2022hatdataset}}} & \multirow{2}{*}{Mean} &&& \multicolumn{4}{c}{\multirowcell{1}[-1.0ex]{HAT~\cite{chung2022hatdataset}}} & \multirow{2}{*}{Mean} & \\
\cmidrule{2-3} \cline{6-8} \cline{10-12} \cline{16-19}
&Action&Scene & & & Places365 & V.C.  & Sin. && Rand. & Close & \multicolumn{1}{c}{Far} & &&& S.O.& Rand. & Close & Far & & \multirow{-4}{*}{H.M.} \\

\midrule

  \multirow{4}{*}{Multi-Task} &\checkmark&\checkmark&  One-Token &\cellcolor{gray!30}91.9 & 10.5 & 5.0 & 21.8 && 19.8 & 27.9 & 8.8 & \cellcolor{gray!30}15.6 & & \cellcolor{gray!30}74.0 & 60.5 & 58.0 & 57.6 & 66.5 & \cellcolor{gray!30}60.7 & \cellcolor{gray!30}38.1   \\
 &\checkmark&\checkmark&  One-Token (unified head) & \cellcolor{gray!30}86.4 & 7.0 & 5.5 & 10.1 && 16.0 & 25.1 & 7.4 & \cellcolor{gray!30}11.9 && \cellcolor{gray!30}73.9 & 60.4 & 57.9 & 58.3 & 69.0 & \cellcolor{gray!30}61.4 & \cellcolor{gray!30}31.9 \\
&\checkmark&\checkmark & Two-Token & \cellcolor{gray!30}86.0 & 11.9 & 11.1 & 19.9 && 17.3 & 23.9 & 9.1 &\cellcolor{gray!30}15.5 && \cellcolor{gray!30}72.3 & 59.6 & 59.2 & 57.8 & 67.1 &\cellcolor{gray!30}60.9 &  \cellcolor{gray!30}37.6 \\
 &\checkmark&\checkmark& Two-Token (unified head) & \cellcolor{gray!30}89.9 & 11.2 & 11.1 & 18.3 && 21.8 & 28.5 & 12.7 &\cellcolor{gray!30}17.3 && \cellcolor{gray!30}74.3 & 61.8 & 61.0 & 59.9 & 69.2 & \cellcolor{gray!30}63.0 & \cellcolor{gray!30}40.7\\

\bottomrule
\end{tabular}

}
\label{tab:unified}
\end{table*}

%% file: table/supple/ablation_full.tex
\begin{table*}[t]
  \centering

  \caption{\textbf{Ablation study.} To validate the effect of each component in the action mask decoder, the cosine similarity loss, backbone architecture variations, and different pre-trained weights, we show the results on the UCF-101 dataset. In every experiment except (c), we use a ViT backbone pre-trained on the UCF-101. We report the Top-1 accuracy (\%) for the action and the Top-5 accuracy (\%) for the scene recognition, along with the harmonic mean (H.M.) of the action and scene recognition performances. The \best{best} H.M. numbers are highlighted.}


    \mpage{0.48}{\scriptsize (a) Effect of cosine similarity loss.}
    \hfill
    \mpage{0.48}{\scriptsize (b) Effect of action mask decoder.}

    \mpage{0.48}{
        \centering
        \resizebox{\linewidth}{!}{
        \begin{tabular}{lccccc}
            \toprule
            \multirow{2}{*}{Method}& \multicolumn{2}{c}{Action($\uparrow$)} & \multicolumn{2}{c}{Scene($\uparrow$)} & \multirow{2}{*}{H.M.}\\
            \cmidrule(lr){2-3} \cmidrule(lr){4-5} 
            
           &  Seen & Unseen & Seen& Unseen \\
            \midrule
            w/o cosine loss  & 90.0 & 42.1 & 70.2 & 57.8 & \cellcolor{gray!30}60.2 \\
           w/ cosine loss  & 90.1 & 40.1  & 74.0 & 61.0 & \cellcolor{gray!30}\best{60.7} \\
           \bottomrule
        \end{tabular}
        }
    }
    \hfill
    \mpage{0.48}{
        \centering
        \resizebox{\linewidth}{!}{
        \begin{tabular}{ccc ccccc}
            \toprule            
            \multicolumn{3}{c}{Method}& \multicolumn{2}{c}{Action($\uparrow$)} & \multicolumn{2}{c}{Scene($\uparrow$)} & \multirow{2}{*}{H.M.}\\
        \cmidrule(lr){1-3} \cmidrule(lr){4-5} \cmidrule(lr){6-7} 

           disen. enc. &Attn. Guidance & Mask pred. & Seen & Unseen & Seen & Unseen \\
            \midrule

            \checkmark&$\times$ &$\times$& 90.0 & 31.6 & 73.7 & 59.8 & \cellcolor{gray!30}54.8 \\
           \checkmark&$\times$  & \checkmark& 89.7 & 33.7 & 73.4 & 60.5& \cellcolor{gray!30}56.4 \\
           \checkmark&\checkmark& $\times$ & 90.8 & 39.9  & 73.8 & 59.8 & \cellcolor{gray!30}60.3 \\

           \checkmark&\checkmark&\checkmark& 90.1 & 40.1 & 74.0 & 61.0 & \cellcolor{gray!30}\best{60.7} \\

           \midrule
           $\times$ &\checkmark&\checkmark& 89.1 & 23.7 & 71.5 & 58.7 & \cellcolor{gray!30}47.3\\
           $\times$ &$\times$&$\times$& 89.5 & 25.3 & 73.2 & 61.4 & \cellcolor{gray!30}49.6\\
           \bottomrule
        \end{tabular}
    
    }
    }

    \mpage{0.48}{\scriptsize (c) Ablations on backbone architecture.}\hfill           
    \mpage{0.48}{\scriptsize (d) Ablation on pre-trained weights.}

    \mpage{0.48}{
        \centering
        \resizebox{0.95\linewidth}{!}{
        \begin{tabular}{lccccc}
            \toprule   
            \multirow{2}{*}{Method} &  \multicolumn{2}{c}{Action($\uparrow$)} & \multicolumn{2}{c}{Scene($\uparrow$)} & \multirow{2}{*}{H.M.}\\
         \cmidrule(lr){2-3} \cmidrule(lr){4-5} 
        & Seen & Unseen & Seen & Unseen \\
        \cmidrule{1-6}
        Naive Action I3D & 92.0 & 20.7 & - & - & \cellcolor{gray!30}- \\
        Naive Scene I3D & - & - & 77.1 & 68.5 & \cellcolor{gray!30}- \\
        \cmidrule{1-6}
        Multi-Task I3D & 90.8 & 20.1 & 72.8 & 56.4 & \cellcolor{gray!30}43.4 \\
        Multi-Task I3D w/ FAME~\cite{fame} & 90.2 & 18.6 & 72.8 & 64.1 & \cellcolor{gray!30}42.5 \\
        \cmidrule{1-6}
        DEVIAS I3D & 91.9 & 29.8 & 76.0 & 67.6 & \cellcolor{gray!30}\textbf{55.3}\\
        \bottomrule
        \end{tabular}
        }
    }
    \hfill    
    \mpage{0.48}{
        \centering
        \resizebox{\linewidth}{!}{
        \begin{tabular}{lccccc}
        \toprule   
        \multirow{2}{*}{Method} &  \multicolumn{2}{c}{Action($\uparrow$)} & \multicolumn{2}{c}{Scene($\uparrow$)} & \multirow{2}{*}{H.M.}\\
         \cmidrule(lr){2-3} \cmidrule(lr){4-5} 
        & Seen & Unseen & Seen & Unseen \\
        \cmidrule{1-6}
        Naive Action Omnivore & 94.1 & 44.0 & - & - & \cellcolor{gray!30}- \\
        Naive Scene Omnivore & - & - & 74.4 & 61.4 & \cellcolor{gray!30}- \\
        \cmidrule{1-6}
        Multi-Task Omnivore & 86.7 & 30.0 & 71.2 & 63.5 & \cellcolor{gray!30}53.6 \\
        Multi-Task Omnivore w/ FAME~\cite{fame}  & 88.8 & 40.1 & 67.1 & 57.1 & \cellcolor{gray!30}58.3 \\
        \cmidrule{1-6}
        DEVIAS Omnivore & 98.6 & 65.3 & 80.5 & 73.0 & \cellcolor{gray!30}\textbf{77.6}\\
       \bottomrule
        \end{tabular}
        }
    }

\label{tab:abl}
\end{table*}

%% file: figure/supple/fig_slot_assignment.tex
\begin{figure}[t]
\centering
\includegraphics[width=.8\linewidth]{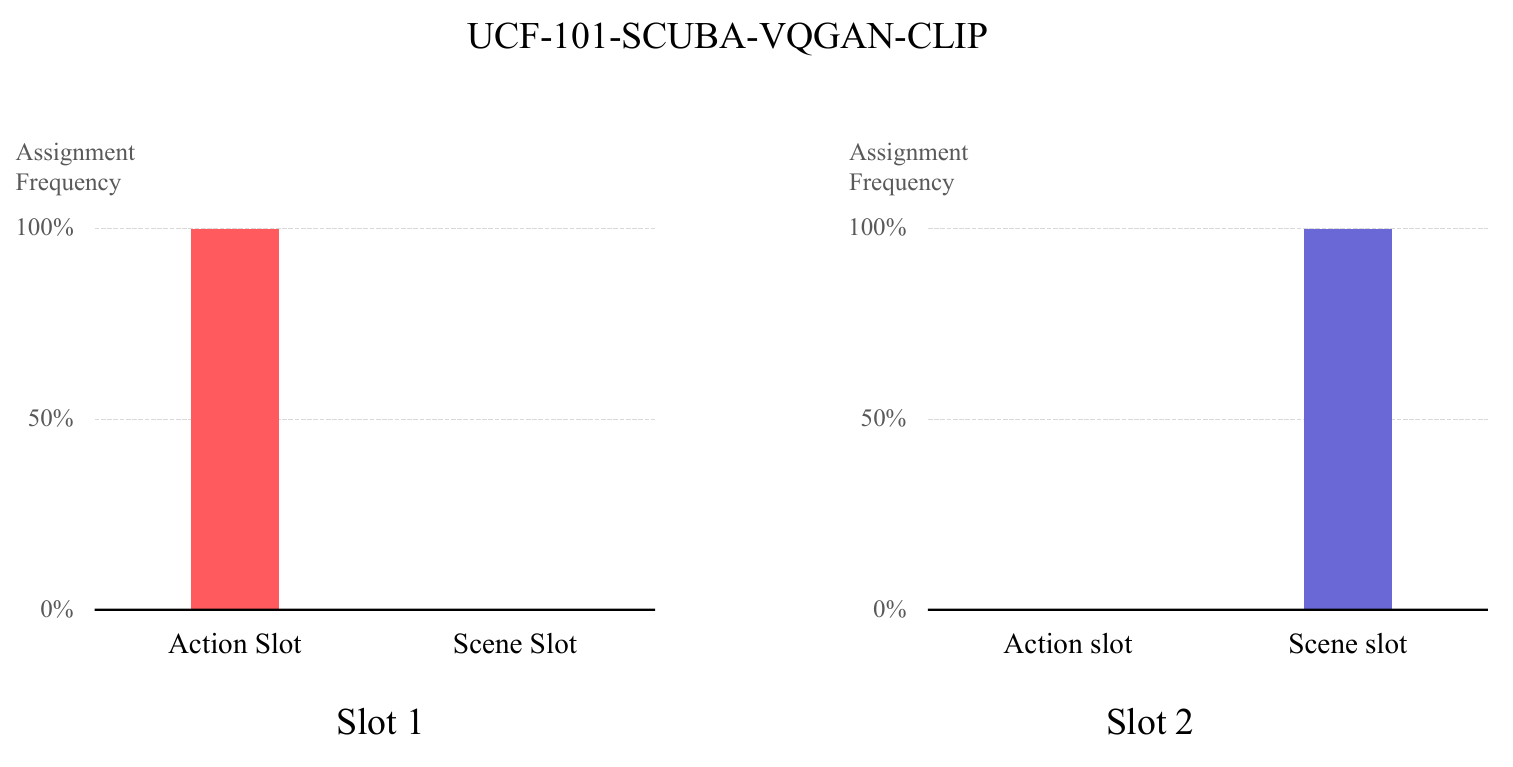}

\caption{\textbf{Slot assignments frequency on UCF-101-SCUBA-VQGAN-CLIP}  We demonstrate the frequency of the slot assignment as either the action slot or the scene slot. In this experiment, we have two slots before assignment ($K=2$): `Slot 1' and `Slot 2'. The result demonstrates that each slot performs a singular role, and the representations of the slots are well-disentangled.}
\label{fig:slot_assign}
\end{figure}

%% file: figure/fig_umap.tex

    

\begin{figure*}[t]
    \mpage{0.20}{
        \centering    
        \includegraphics[width=\linewidth]{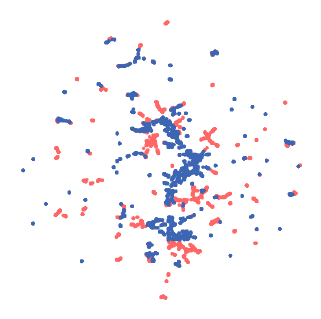}
    }    
    \mpage{0.20}{
        \centering
        \includegraphics[width=\linewidth]{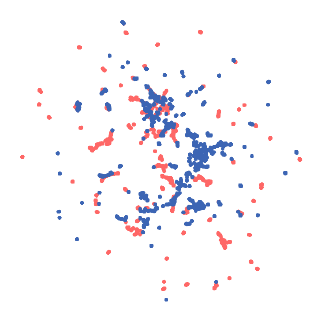}
    }    
    \mpage{0.20}{
        \centering
        \includegraphics[width=\linewidth]{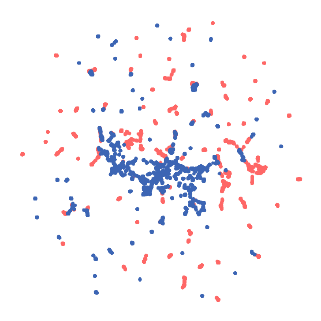}
    }        
    \mpage{0.20}{
        \centering
        \includegraphics[width=\linewidth]{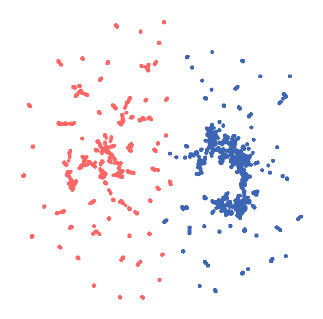}
    }    
    \mpage{0.10}{
        \centering
        \includegraphics[width=\linewidth]{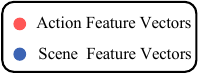}
    }
    
    \mpage{0.20}{\fontsize{7pt}{10pt}\selectfont (a) Two-Token.}
    \mpage{0.20}{\fontsize{7pt}{10pt}\selectfont (b) Two-Token w/ BE~\cite{BE}.}    
    \mpage{0.20}{\fontsize{7pt}{10pt}\selectfont (c) Two-Token w/ FAME~\cite{fame}.}
    \mpage{0.20}{\fontsize{7pt}{10pt}\selectfont (d) DEVIAS.}
    \mpage{0.10}{}    

    \captionof{figure}{
    \textbf{UMAP~\cite{mcinnes2018umap} visualization of feature vectors on UCF-101 dataset, test split.} (a) Two-Token, (b) Two-Token w/ BE, and (c) Two-Token w/ FAME baselines show entangled action and scene feature vectors. In contrast, (d) our DEVIAS demonstrates a distinct separation between action and scene feature vectors. All models are trained on the UCF-101 train split. Best viewed with zoom and color.
    }
    \label{fig:umap}
\end{figure*}

%% file: figure/supple/rebuttal/fig_attnmap.tex
\begin{figure*}[t]
    \centering
    \mpage{0.31}{
        \includegraphics[width=0.8\linewidth]{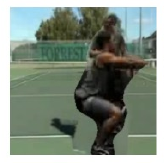}
    }    
    \hfill
    \mpage{0.31}{
        \includegraphics[width=0.8\linewidth]{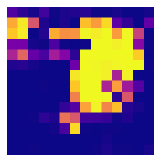}
    }
    \hfill
    \mpage{0.31}{
        \includegraphics[width=0.8\linewidth]{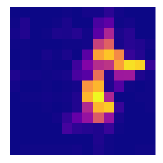}
    }

    \mpage{0.31}{\footnotesize (a) Input frame.}
    \mpage{0.31}{\footnotesize (b) without AMD.}    
    \mpage{0.31}{\footnotesize (c) with AMD.}
    \caption{\textbf{Effect of action mask decoder (AMD). } 
    a) When training DEVIAS without AMD, the action slot attention is a bit noisy \ie attending to scene regions as well. (b) When training DEVIAS with AMD, the action slot more accurately focuses on the action region. 
    The result implies AMD is effective in learning disentangled action and scene representations.
     }
    \label{fig:attnmap}
\end{figure*}

%% file: figure/supple/fig_full_slot_viz.tex
\begin{figure*}[h]


    \mpage{0.08}{\footnotesize Input 
    Frame
    }    
    \hfill
    \mpage{0.38}{
        \centering
        \includegraphics[width=\linewidth]{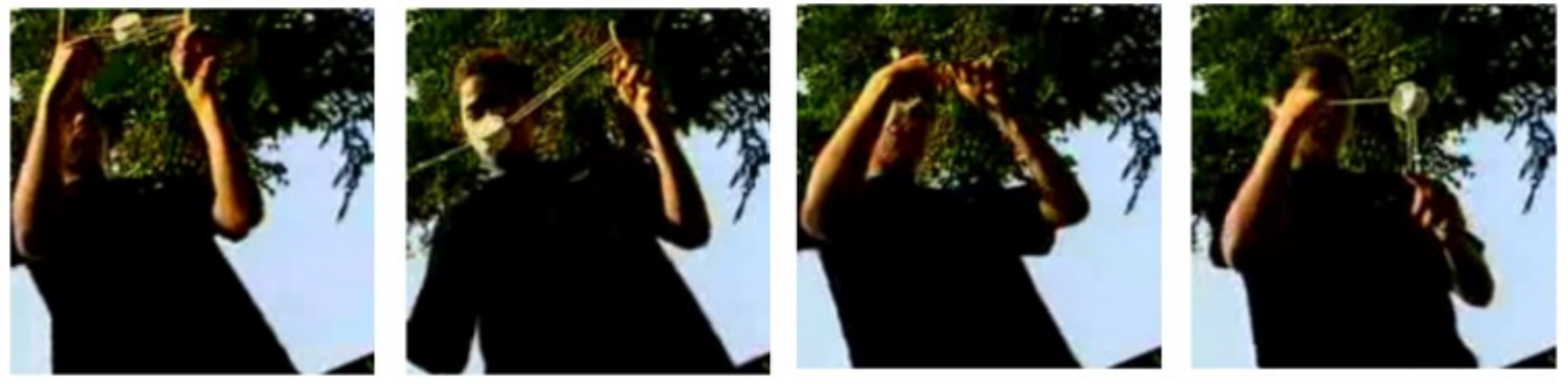}
    }
    \hfill
    \mpage{0.38}{
        \centering
        \includegraphics[width=\linewidth]{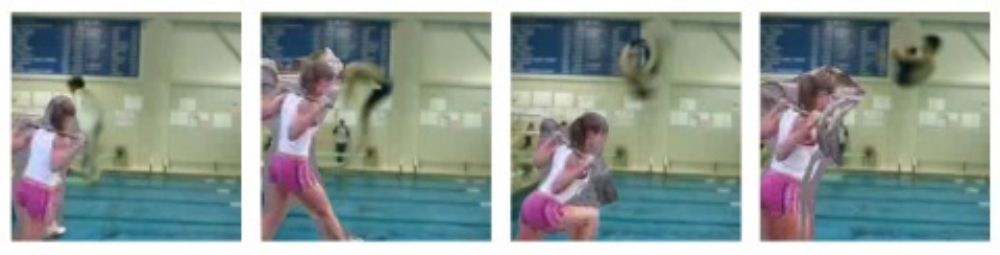}
    }
    
    \mpage{0.08}{
    \footnotesize Action
    Slot
    } 
    \hfill
    \mpage{0.38}{
        \centering
        \includegraphics[width=\linewidth]{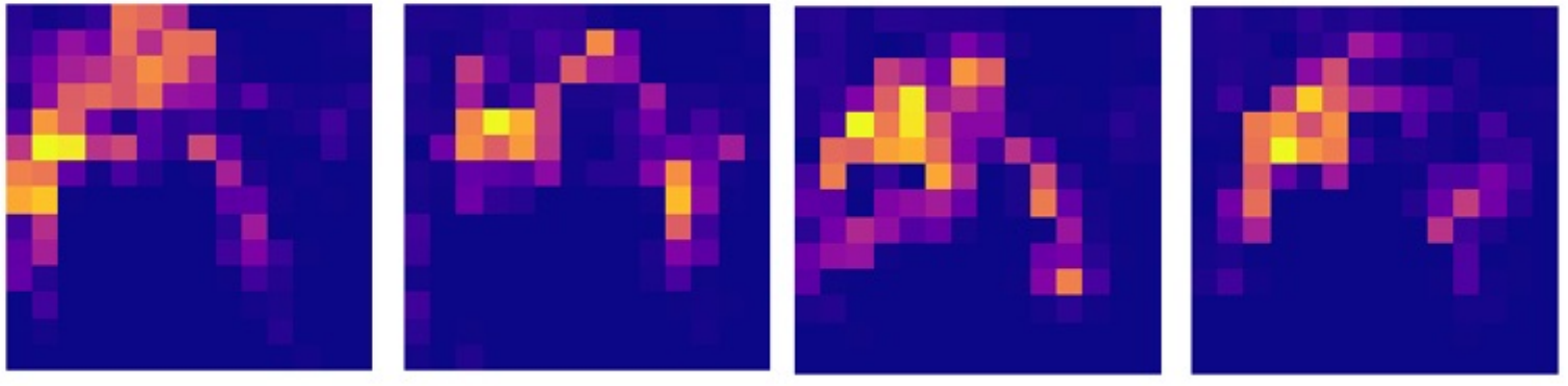}
    }
    \hfill
    \mpage{0.38}{
        \centering
        \includegraphics[width=\linewidth]{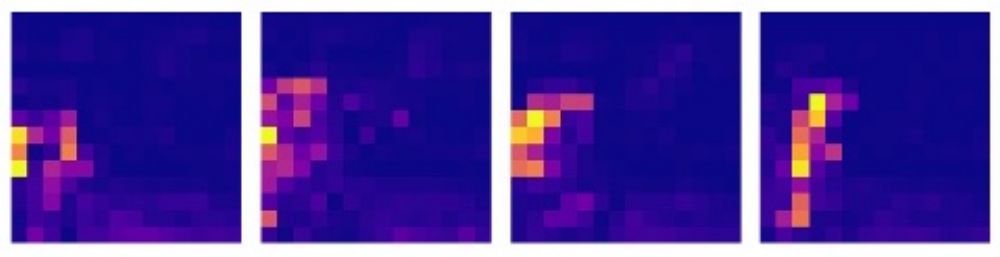}
    }
    
    \mpage{0.08}{
    \footnotesize Scene
    Slot
    }   
    \hfill
    \mpage{0.38}{
        \centering
        \includegraphics[width=\linewidth]{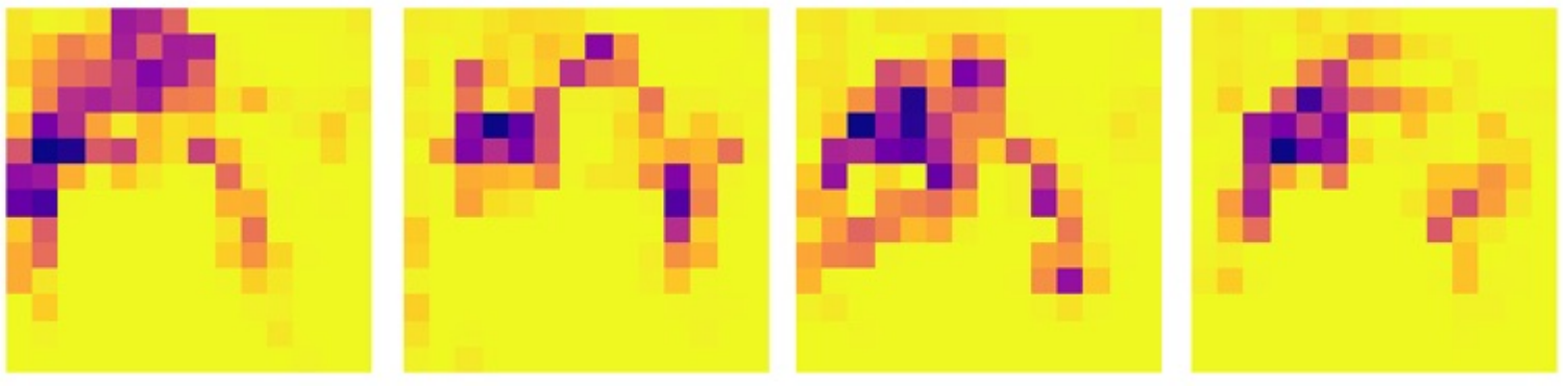}
    }
    \hfill
    \mpage{0.38}{
        \centering
        \includegraphics[width=\linewidth]{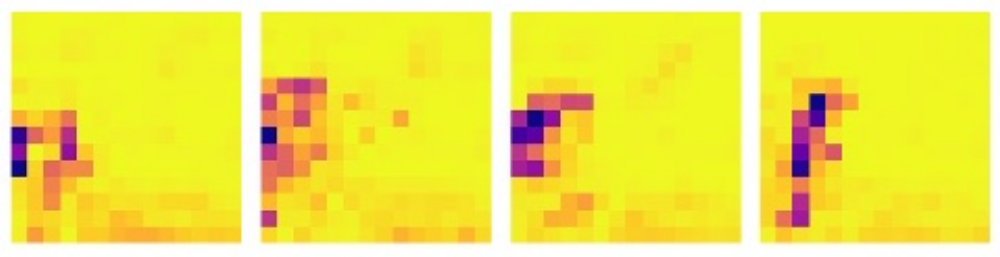}
    }


    \mpage{0.08}{\footnotesize Input 
    Frame
    }    
    \hfill
    \mpage{0.38}{
        \centering
        \includegraphics[width=\linewidth]{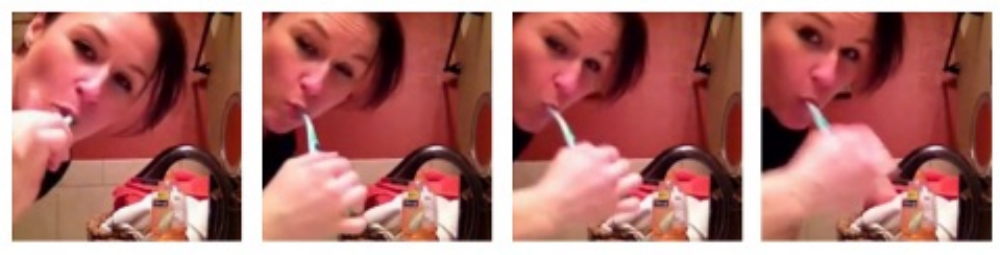}
    }
    \hfill
    \mpage{0.38}{
        \centering
        \includegraphics[width=\linewidth]{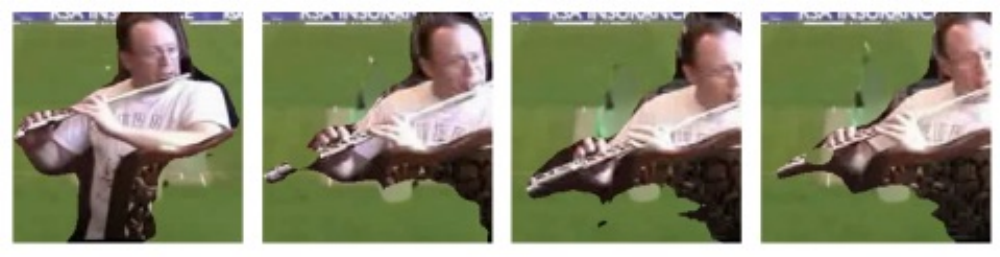}
    }
    
    \mpage{0.08}{
    \footnotesize Action
    Slot
    } 
    \hfill
    \mpage{0.38}{
        \centering
        \includegraphics[width=\linewidth]{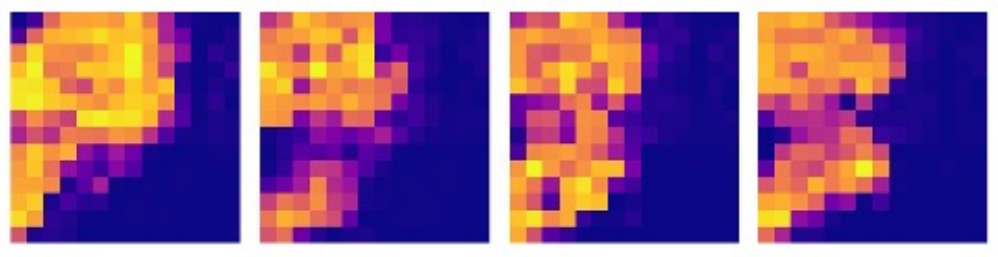}
    }
    \hfill
    \mpage{0.38}{
        \centering
        \includegraphics[width=\linewidth]{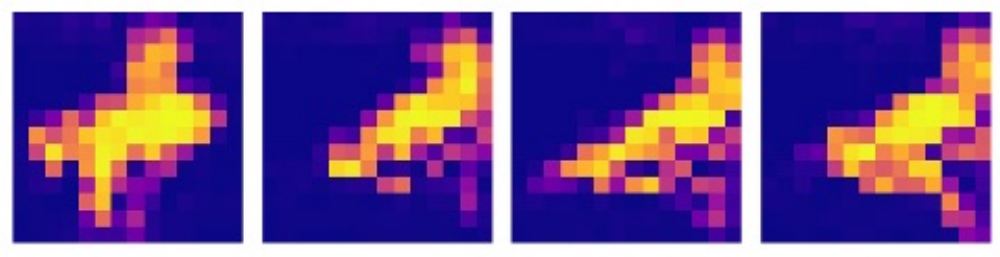}
    }
    
    \mpage{0.08}{
    \footnotesize Scene
    Slot
    }   
    \hfill
    \mpage{0.38}{
        \centering
        \includegraphics[width=\linewidth]{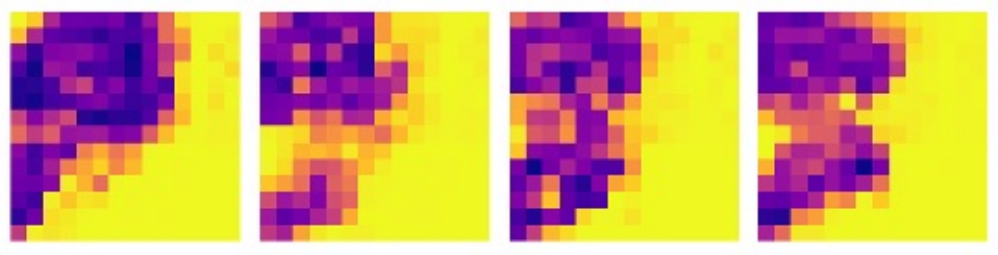}
    }
    \hfill
    \mpage{0.38}{
        \centering
        \includegraphics[width=\linewidth]{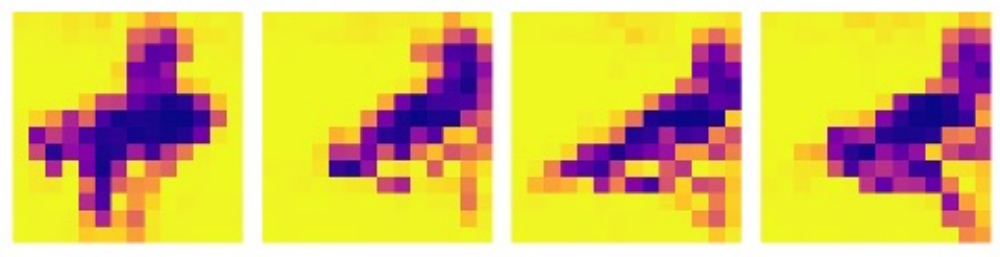}
    }

    
    \mpage{0.08}{\footnotesize Input 
    Frame
    }    
    \hfill
    \mpage{0.38}{
        \centering
        \includegraphics[width=\linewidth]{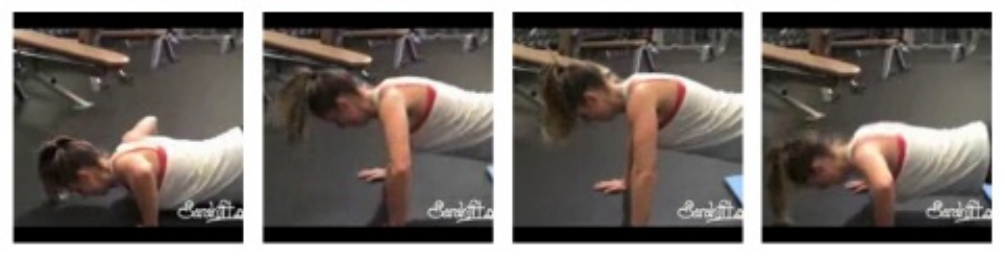}
    }
    \hfill
    \mpage{0.38}{
        \centering
        \includegraphics[width=\linewidth]{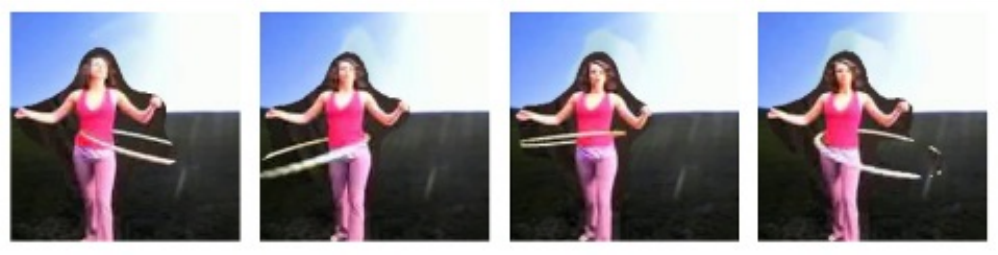}
    }
    
    \mpage{0.08}{
    \footnotesize Action
    Slot
    } 
    \hfill
    \mpage{0.38}{
        \centering
        \includegraphics[width=\linewidth]{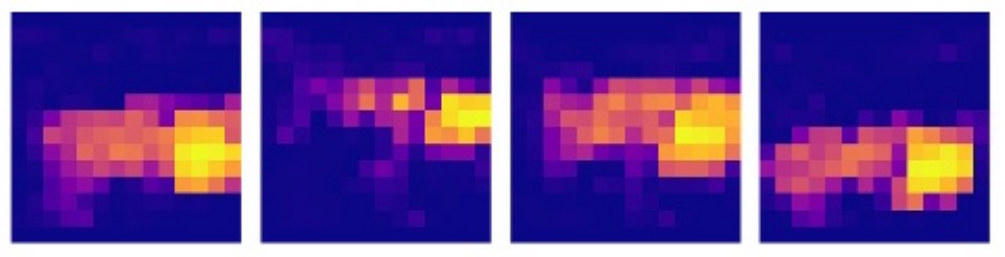}
    }
    \hfill
    \mpage{0.38}{
        \centering
        \includegraphics[width=\linewidth]{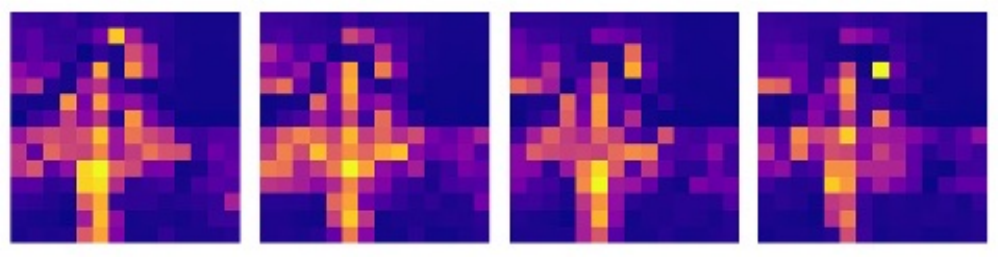}
    }
    
    \mpage{0.08}{
    \footnotesize Scene
    Slot
    }   
    \hfill
    \mpage{0.38}{
        \centering
        \includegraphics[width=\linewidth]{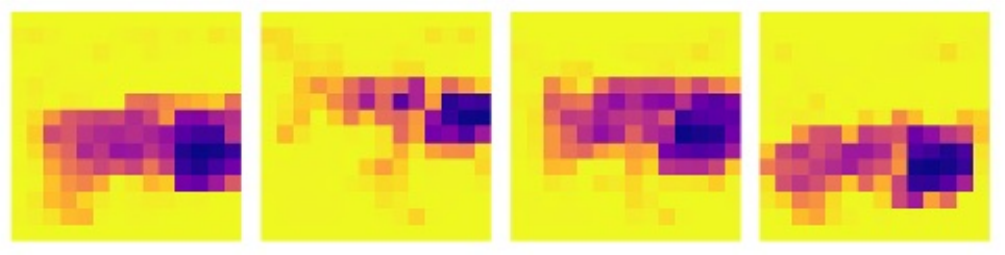}
    }
    \hfill
    \mpage{0.38}{
        \centering
        \includegraphics[width=\linewidth]{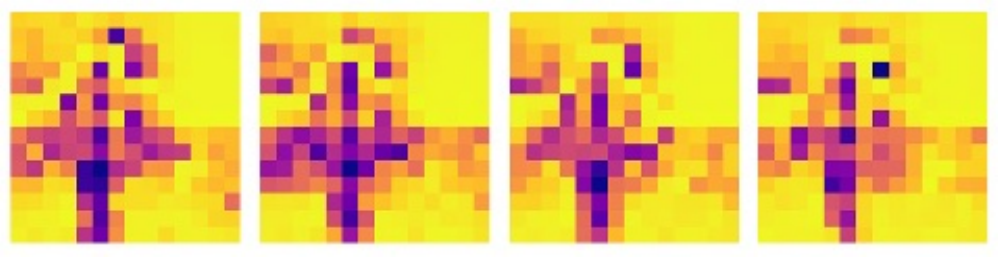}
    }

    
    \mpage{0.08}{\footnotesize Input 
    Frame
    }    
    \hfill
    \mpage{0.38}{
        \centering
        \includegraphics[width=\linewidth]{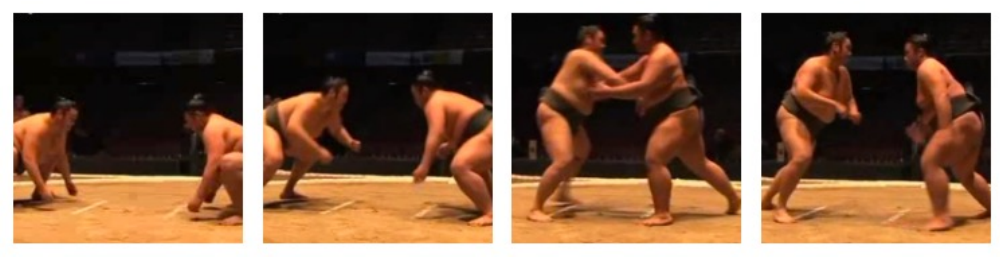}
    }
    \hfill
    \mpage{0.38}{
        \centering
        \includegraphics[width=\linewidth]{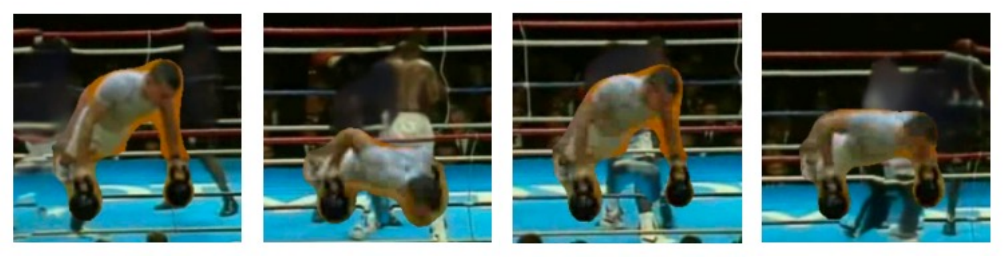}
    }
    
    \mpage{0.08}{
    \footnotesize Action
    Slot
    } 
    \hfill
    \mpage{0.38}{
        \centering
        \includegraphics[width=\linewidth]{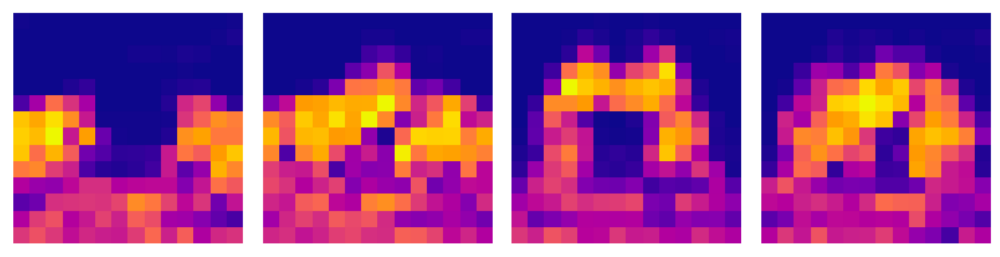}
    }
    \hfill
    \mpage{0.38}{
        \centering
        \includegraphics[width=\linewidth]{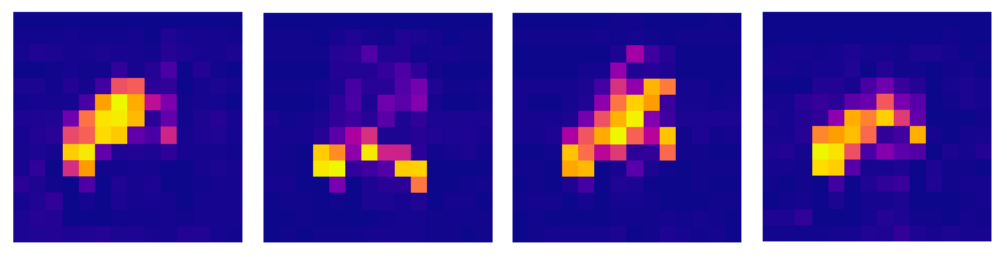}
    }
    
    \mpage{0.08}{
    \footnotesize Scene
    Slot
    }   
    \hfill
    \mpage{0.38}{
        \centering
        \includegraphics[width=\linewidth]{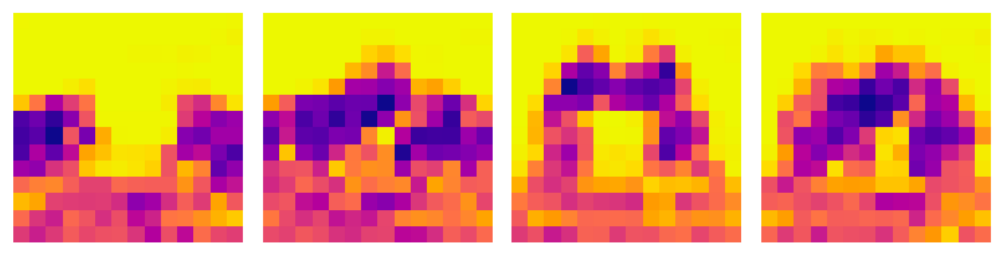}
    }
    \hfill
    \mpage{0.38}{
        \centering
        \includegraphics[width=\linewidth]{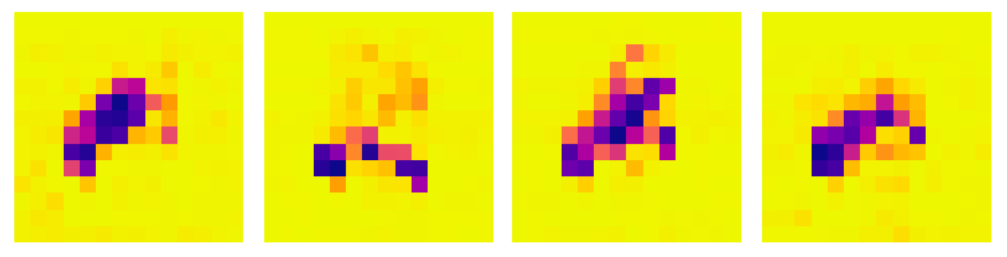}
    }


    \mpage{0.09}{}   \hfill
    \mpage{0.42}{\footnotesize (a) Seen combinations}
    \mpage{0.42}{\footnotesize (b) Unseen combinations}
    \\

    \captionof{figure}{
        \textbf{Visualization of DEVIAS slot attention map.}
        (a) Seen combinations video from UCF-101~\cite{soomro2012ucf101}, (b) Unseen combinations video from HAT Far~\cite{chung2022hatdataset}. 
        Each slot attends to action and scene regions well across frames regardless of seen or unseen combinations. Best viewed with zoom and color.
    }
    \label{fig:full_slot_viz}
\end{figure*}

%% file: table/supple/scene_hyper.tex
\begin{table}[p]
\centering
\captionsetup{font=scriptsize}
\caption{\tb{Hyperparameters used for training the scene model on Places365.} 
}
\resizebox{.47\columnwidth}{!}{
\begin{tabular}{lc}
    
\toprule
Config & Places365\\
\midrule
Optimizer & AdamW~\cite{adamw} \\
Base learning rate & 5e-4 \\
Weight decay & 0.05   \\
Optimizer momentum & {$\beta_1, \beta_2 = 0.9, 0.999$}~\cite{chen2020generative} \\
Per GPU batch size &  32 \\
Drop path& 0.1  \\
Mixup~\cite{zhang2017mixup}& 0.8  \\
Cutmix~\cite{yun2019cutmix}& 1.0  \\
Smoothing~\cite{labelsmo}& 0.1  \\
Flip augmentation & \checkmark  \\
Update frequency  & 2 \\
Learning rate schedule  & cosine decay~\cite{cosineannealing} \\
Warmup epochs & 5 \\
Layer-wise learning rate decay&  0.65  \\
Training epochs &  100  \\
\bottomrule
\end{tabular}
}
\label{tab:scene_hyper}
\end{table}

%% file: table/supple/action_noaug_hyper.tex
\begin{table}[t]
\centering
\captionsetup{font=scriptsize}
\caption{\tb{Hyperparameters used for training the Naive Action ViT, One-Token, and Two-Token baselines on each dataset.}}

\resizebox{0.8\columnwidth}{!}{
\begin{tabular}{lccc}
    
\toprule
Config & UCF-101 & Kinetics-400& HVU \\
\midrule
Optimizer &   AdamW~\cite{adamw}& AdamW& AdamW \\
Base learning rate &   5e-4 &1e-3 &   5e-4\\
Weight decay &   0.05 &0.05 &0.05  \\
Optimizer momentum &   {$\beta_1, \beta_2 = 0.9, 0.999$}~\cite{chen2020generative}& {$\beta_1, \beta_2 = 0.9, 0.999$} & {$\beta_1, \beta_2 = 0.9, 0.999$} \\
Per GPU batch size &    12&12&12 \\
Drop out & 0.5 & 0.0& 0.0\\
Drop path& 0.2 & 0.1 & 0.1 \\
Color jitter&   0.4&0.4 &0.4 \\
Flip augmentation & \checkmark & \checkmark & \checkmark \\
Mixup~\cite{zhang2017mixup} & $\times$ & $\times$ & $\times$ \\
Cutmix~\cite{yun2019cutmix} & $\times$ & $\times$ & $\times$ \\
Random erasing~\cite{zhong2020random} & $\times$ & $\times$& $\times$\\
Layer-wise learning rate decay~\cite{bao2021beit}&    0.75&0.75&0.75 \\
Learning rate schedule  & cosine decay~\cite{cosineannealing} & cosine decay & cosine decay\\
Warmup epochs & 5 & 5& 5 \\
Training epochs &    100&100&50 \\
\bottomrule
\end{tabular}
}

\label{tab:action_noaug_hyper}
\end{table}

%% file: table/supple/action_aug_hyper.tex
\begin{table}[t]
\centering
\captionsetup{font=scriptsize}

\caption{\tb{Hyperparameters used for training the Action ViT w/ Aug. on each dataset.} 
}

\resizebox{.67\columnwidth}{!}{
\begin{tabular}{lcc}
    
\toprule
Config & UCF-101 & Kinetics-400\\
\midrule
Optimizer &   AdamW~\cite{adamw}& AdamW\\
Base learning rate &   5e-4 &1e-3 \\
Weight decay &   0.05 &0.05   \\
Optimizer momentum &   {$\beta_1, \beta_2 = 0.9, 0.999$}~\cite{chen2020generative}& {$\beta_1, \beta_2 = 0.9, 0.999$} \\
Per GPU batch size &    12&12 \\
Drop out & 0.5 & 0.0\\
Drop path& 0.2 & 0.1  \\
Color jitter&   0.4&0.4 \\
Flip augmentation & \checkmark & \checkmark \\
Mixup~\cite{zhang2017mixup} & 0.8 & 0.8 \\
Cutmix~\cite{yun2019cutmix} & 1.0 & 1.0 \\
Random erasing~\cite{zhong2020random} & 0.25 & 0.25\\
Layer-wise learning rate decay~\cite{bao2021beit}&    0.75&0.75 \\
Learning rate schedule  & cosine decay~\cite{cosineannealing} & cosine decay\\
Warmup epochs & 5 & 5 \\
Training epochs &    100&100 \\
\bottomrule
\end{tabular}
}

\label{tab:action_aug_hyper}
\end{table}

%% file: table/supple/scene_vit_hyper.tex
\begin{table}[t]
\centering
\captionsetup{font=scriptsize}
\caption{
\tb{Hyperparameters used for training the Naive Scene ViT on each dataset.} 
}

\resizebox{.78\columnwidth}{!}{
\begin{tabular}{l ccc}
    
\toprule
Config & UCF-101 & Kinetics-400& HVU \\
\midrule
Optimizer &   AdamW~\cite{adamw}& AdamW & AdamW \\
Base learning rate &   5e-4 &1e-3 &   5e-4 \\
Weight decay &   0.05 &0.05 &0.05  \\
Optimizer momentum &   {$\beta_1, \beta_2 = 0.9, 0.999$}~\cite{chen2020generative}& {$\beta_1, \beta_2 = 0.9, 0.999$} & {$\beta_1, \beta_2 = 0.9, 0.999$} \\
Per GPU batch size &    12&12 &12 \\
Drop path& 0.2 & 0.1& 0.1 \\
Color jitter&   0.4&0.4&0.4 \\
Flip augmentation & \checkmark & \checkmark& \checkmark \\
Mixup~\cite{zhang2017mixup} & $\times$ & $\times$ & $\times$ \\
Cutmix~\cite{yun2019cutmix} & $\times$ & $\times$&$\times$ \\
Random erasing~\cite{zhong2020random} & $\times$ & $\times$&$\times$\\
Layer-wise learning rate decay~\cite{bao2021beit}&    0.75&0.75 &0.75 \\
Learning rate schedule  & cosine decay~\cite{cosineannealing} & cosine decay& cosine decay\\
Warmup epochs & 5 & 5& 5 \\
Training epochs &    100&100&50 \\
\bottomrule
\end{tabular}
}

\label{tab:scene_vit_hyper}
\end{table}

%% file: table/supple/be.tex
\begin{table}[t]
\centering
\captionsetup{font=scriptsize}
\caption{\tb{Hyperparameters used for BE~\cite{BE} and training the Two-Token w/ BE on each dataset.}}

\resizebox{.62\columnwidth}{!}{
\begin{tabular}{lcc}
    
\toprule
Config & UCF-101 & Kinetics-400 \\
\midrule
Optimizer &   AdamW~\cite{adamw}& AdamW \\
Base learning rate &   5e-4 &1e-3 \\
Weight decay &   0.05 &0.05  \\
Optimizer momentum &   {$\beta_1, \beta_2 = 0.9, 0.999$}~\cite{chen2020generative}& {$\beta_1, \beta_2 = 0.9, 0.999$} \\
Per GPU batch size &    12&12 \\
Drop out & 0.5 & 0.0\\
Drop path& 0.2 & 0.1 \\
Color jitter&   0.4&0.4 \\
Flip augmentation & \checkmark & \checkmark \\
Mixup~\cite{zhang2017mixup} & $\times$ & $\times$ \\
Cutmix~\cite{yun2019cutmix} & $\times$ & $\times$ \\
Random erasing~\cite{zhong2020random} & $\times$ & $\times$\\
Mixing weight & $\sim \text{Uniform}(0,0.3)$ & $\sim \text{Uniform}(0,0.3)$\\
$\rho$ & 0.5 & 0.5 \\
Layer-wise learning rate decay~\cite{bao2021beit}&    0.75&0.75 \\
Learning rate schedule  & cosine decay~\cite{cosineannealing} & cosine decay\\
Warmup epochs & 5 & 5 \\
Training epochs &    100&100 \\
\bottomrule
\end{tabular}
}

\label{tab:be_hyper}
\end{table}

%% file: table/supple/fame.tex
\begin{table}[t]
\centering
\captionsetup{font=scriptsize}
\caption{\tb{Hyperparameters used for FAME~\cite{fame} and training the Two-Token w/ FAME on each dataset.}}

\resizebox{.78\columnwidth}{!}{
\begin{tabular}{lccc}
    
\toprule
Config & UCF-101 & Kinetics-400 & HVU\\
\midrule
Optimizer &   AdamW~\cite{adamw}& AdamW & AdamW \\
Base learning rate &   5e-4 &1e-3  &   5e-4 \\
Weight decay &   0.05 &0.05  &0.05  \\
Optimizer momentum &   {$\beta_1, \beta_2 = 0.9, 0.999$}~\cite{chen2020generative}& {$\beta_1, \beta_2 = 0.9, 0.999$} & {$\beta_1, \beta_2 = 0.9, 0.999$} \\
Per GPU batch size &    12&12&12 \\
Drop out & 0.5 & 0.0& 0.0\\
Drop path& 0.2 & 0.1 & 0.1 \\
Color jitter&   0.4&0.4&0.4 \\
Flip augmentation & \checkmark & \checkmark & \checkmark \\
Mixup~\cite{zhang2017mixup} & $\times$ & $\times$ & $\times$ \\
Cutmix~\cite{yun2019cutmix} & $\times$ & $\times$& $\times$ \\
Random erasing~\cite{zhong2020random} & $\times$ & $\times$& $\times$\\
$\tau$,$\rho$ & 0.5,0.5 & 0.5,0.5& 0.5,0.5 \\
Layer-wise learning rate decay~\cite{bao2021beit}&    0.75&0.75&0.75 \\
Learning rate schedule  & cosine decay~\cite{cosineannealing} & cosine decay& cosine decay\\
Warmup epochs & 5 & 5& 5 \\
Training epochs &    100&100&50 \\
\bottomrule
\end{tabular}
}

\label{tab:fame_hyper}
\end{table}

%% file: table/supple/linearprobe_hyper.tex
\begin{table}[t]
\centering
\captionsetup{font=scriptsize}
\caption{\tb{Hyperparameters used for the linear probe experiments.} 
}
\resizebox{0.7\columnwidth}{!}{

\begin{tabular}{lccc}
    
\toprule
Config & UCF-101 & Kinetics-400 & HVU\\
\midrule
Optimizer &  AdamW~\cite{adamw}&AdamW&AdamW\\
Base learning rate &  1e-3&1e-3&5e-4\\
Weight decay &  0.05&0.05&0.05 \\
Optimizer momentum &  $\beta_1, \beta_2 = 0.9, 0.999$~\cite{chen2020generative}& $\beta_1, \beta_2 = 0.9, 0.999$& $\beta_1, \beta_2 = 0.9, 0.999$\\
Per GPU batch size &  24 & 16& 16 \\
Color jitter&  0.4&0.4&0.4 \\
Flip augmentation & \checkmark & \checkmark & \checkmark \\
Mixup~\cite{zhang2017mixup} & $\times$ & $\times$& $\times$ \\
Cutmix~\cite{yun2019cutmix} & $\times$ & $\times$ & $\times$\\
Random erasing~\cite{zhong2020random} & $\times$ & $\times$ & $\times$ \\
Learning rate schedule  & cosine decay~\cite{cosineannealing} & cosine decay& cosine decay\\
Warmup epochs & 5 & 5& 5 \\
Training epochs & 50&50&50 \\
\bottomrule
\end{tabular}
}

\label{tab:lp_hyper}
\end{table}

%% file: table/supple/hyperpramas.tex
\begin{table}[t]
\centering
\captionsetup{font=scriptsize}
\caption{\tb{Hyperparameters used for training DEVIAS on each dataset.} 
}
\resizebox{.78\columnwidth}{!}{

\begin{tabular}{lccc}
    
\toprule
Config & UCF-101 & Kinetics-400& HVU \\
\midrule
Optimizer &   AdamW~\cite{adamw}& AdamW& AdamW \\
Base learning rate &   5e-4 &5e-4 &5e-4 \\
Weight decay &   0.05 &0.05  \\
Optimizer momentum &   {$\beta_1, \beta_2 = 0.9, 0.999$}~\cite{chen2020generative}& {$\beta_1, \beta_2 = 0.9, 0.999$}& {$\beta_1, \beta_2 = 0.9, 0.999$} \\
Per GPU batch size &    12&12&12 \\
Drop out & 0.5 & 0.0& 0.0\\
Drop path& 0.2 & 0.1 & 0.1 \\
Color jitter&   0.4&0.4 &0.4 \\
Flip augmentation & \checkmark & \checkmark & \checkmark \\
Mixup~\cite{zhang2017mixup} & $\times$ & $\times$  & $\times$ \\
Cutmix~\cite{yun2019cutmix} & $\times$ & $\times$& $\times$ \\
Random erasing~\cite{zhong2020random} & $\times$ & $\times$& $\times$ \\
Layer-wise learning rate decay~\cite{bao2021beit}&    0.75&0.75&0.75 \\
Learning rate schedule  & cosine decay~\cite{cosineannealing} & cosine decay& cosine decay\\
Warmup epochs & 5 & 5& 5 \\
Mask extractor &   FAME~\cite{fame}&FAME&FAME \\
$\tau$,$\rho$ & 0.3,0.4 & 0.5,0.8& 0.5,0.25 \\
$\alpha,\beta,\gamma$ &   1,1,1&1,1,1&1,1,1 \\
M & 4 & 8& 8 \\
Slot attention learning scale &   0.1&0.1&0.1  \\
K &  2&2&2 \\
Training epochs &    100&100&50 \\
\bottomrule
\end{tabular}
}

\label{tab:hyper}
\end{table}

%% file: table/supple/downstream_hyper.tex
\begin{table}[t]
\centering
\captionsetup{font=scriptsize}
\caption{\tb{Hyperparameters used for the downstream task fine-tuning on each target dataset.}}

\resizebox{.88\columnwidth}{!}{
\begin{tabular}{lcccc}
     
\toprule
Config &Diving48&SSV2& UCF-101 & ActivityNet \\
\midrule
Optimizer &   AdamW~\cite{adamw}& AdamW& AdamW & AdamW \\
Base learning rate &   5e-4 & 5e-4 & 5e-4 & 5e-4\\
Weight decay &   0.05 &0.05 &0.05 &0.05  \\
Optimizer momentum &   {$\beta_1, \beta_2 = 0.9, 0.999$}~\cite{chen2020generative}& {$\beta_1, \beta_2 = 0.9, 0.999$} & {$\beta_1, \beta_2 = 0.9, 0.999$} & {$\beta_1, \beta_2 = 0.9, 0.999$} \\
Per GPU batch size &    12&12&12 &12 \\
Drop path& 0.1 & 0.1 & 0.1 & 0.1 \\
Color jitter&   0.4&0.4 &0.4 &0.4 \\
Flip augmentation & \checkmark & $\times$ & \checkmark & \checkmark \\
Mixup~\cite{zhang2017mixup} & $\times$ & $\times$ &$\times$ &$\times$ \\
Cutmix~\cite{yun2019cutmix} & $\times$ & $\times$ &$\times$ &$\times$\\
Random erasing~\cite{zhong2020random} & $\times$ & $\times$&$\times$ &$\times$\\
Layer-wise learning rate decay~\cite{bao2021beit}&    0.75&0.75&0.75 & 0.75 \\
Learning rate schedule  & cosine decay~\cite{cosineannealing} & cosine decay & cosine decay & cosine decay\\
Warmup epochs & 5 & 5& 5 & 5 \\
Training epochs &    50&50&50 &50 \\
\bottomrule
\end{tabular}
}

\label{tab:downstream_hyper}
\end{table}